\documentclass[conference]{IEEEtran}
\usepackage{times}

\usepackage[numbers]{natbib}
\usepackage{multicol}
\usepackage[bookmarks=true]{hyperref}
\usepackage{url}
\usepackage{algorithm}
\usepackage{algorithmic}
\usepackage{graphicx}
\usepackage{diagbox}
\usepackage{multirow}
\usepackage{subfigure}
\usepackage{comment}
\usepackage{placeins}
\usepackage{dblfloatfix}
\usepackage[table]{xcolor}
\usepackage[bottom]{footmisc}

\usepackage{amsmath,amsfonts,bm}

\def\eqref#1{equation~\ref{#1}}

\def\1{\bm{1}}

\def\rva{{\mathbf{a}}}

\def\rvg{{\mathbf{g}}}

\def\rvm{{\mathbf{m}}}

\def\rvq{{\mathbf{q}}}

\def\rvs{{\mathbf{s}}}

\def\rvx{{\mathbf{x}}}

\def\rvz{{\mathbf{z}}}

\def\rvI{{\mathbf{I}}}

\def\rvW{{\mathbf{W}}}
\def\rvZ{{\mathbf{Z}}}
\def\rvM{{\mathbf{M}}}
\def\rvSigma{{\mathbf{\Sigma}}}
\def\rvmu{{\text{\boldmath$\mu$}}}

\DeclareMathAlphabet{\mathsfit}{\encodingdefault}{\sfdefault}{m}{sl}
\SetMathAlphabet{\mathsfit}{bold}{\encodingdefault}{\sfdefault}{bx}{n}

\newcommand{\expec}{\mathbb{E}}

\begin{document}

\title{Learning Agile Robotic Locomotion Skills by Imitating Animals}


\author{
    \IEEEauthorblockN{Xue Bin Peng\IEEEauthorrefmark{1}\IEEEauthorrefmark{2}, Erwin Coumans\IEEEauthorrefmark{1}, Tingnan Zhang\IEEEauthorrefmark{1}, Tsang-Wei Edward Lee\IEEEauthorrefmark{1}, Jie Tan\IEEEauthorrefmark{1}, Sergey Levine\IEEEauthorrefmark{1}\IEEEauthorrefmark{2}}
    \IEEEauthorblockA{\IEEEauthorrefmark{1}Google Research, \qquad \IEEEauthorrefmark{2}University of California, Berkeley}
    \IEEEauthorblockA{Email: xbpeng@berkeley.edu, \{erwincoumans,tingnan,tsangwei,jietan\}@google.com, svlevine@eecs.berkeley.edu}
}


%

\maketitle

\begin{figure}[t]
\begin{minipage}{1\textwidth}
    \includegraphics[width=1\textwidth]{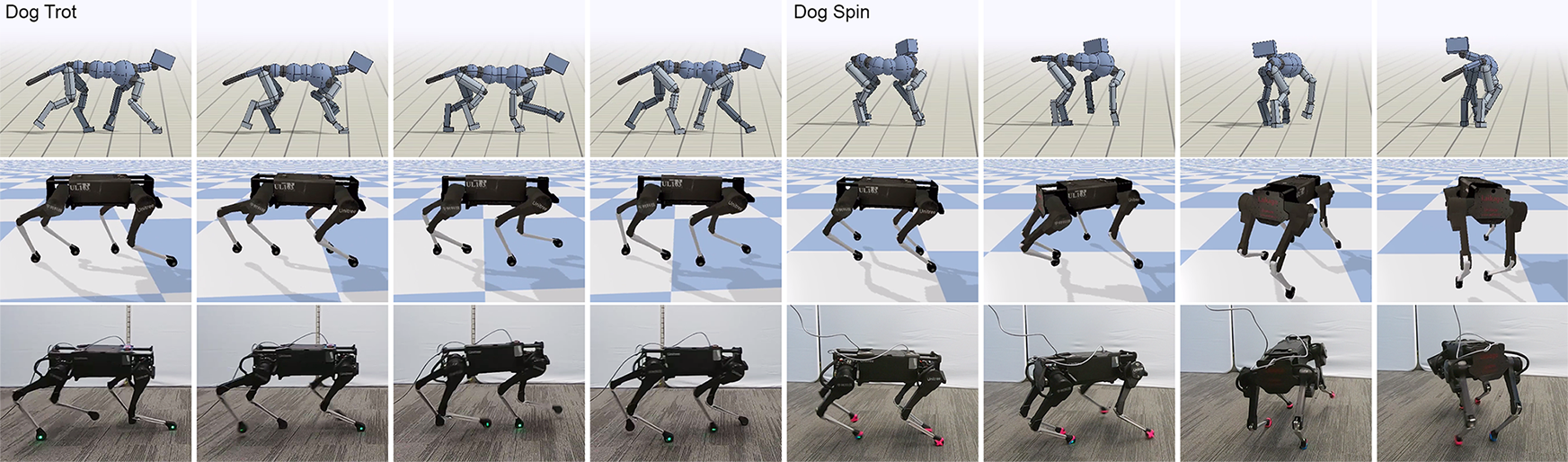}
    \vspace{-0.5cm}
\caption{Laikago robot performing locomotion skills learned by imitating motion data recorded from a real dog. \textbf{Top:} Motion capture data recorded from a dog. \textbf{Middle:} Simulated Laikago robot imitating reference motions.
\textbf{Bottom:} Real Laikago robot imitating reference motions.}
\label{fig:teaser}
\end{minipage}
\vspace{-0.5cm}
\end{figure}

\begin{abstract}
Reproducing the diverse and agile locomotion skills of animals has been a longstanding challenge in robotics. While manually-designed controllers have been able to emulate many complex behaviors, building such controllers involves a time-consuming and difficult development process, often requiring substantial  expertise of the nuances of each skill. Reinforcement learning provides an appealing alternative for automating the manual effort involved in the development of controllers. However, designing learning objectives that elicit the desired behaviors from an agent can also require a great deal of skill-specific expertise. In this work, we present an imitation learning system that enables legged robots to learn agile locomotion skills by imitating real-world animals. We show that by leveraging reference motion data, a single learning-based approach is able to automatically synthesize controllers for a diverse repertoire behaviors for legged robots. By incorporating sample efficient domain adaptation techniques into the training process, our system is able to learn adaptive policies in simulation that can then be quickly adapted for real-world deployment. To demonstrate the effectiveness of our system, we train an 18-DoF quadruped robot to perform a variety of agile behaviors ranging from different locomotion gaits to dynamic hops and turns. (\href{https://xbpeng.github.io/projects/Robotic_Imitation/}{Video\footnote{\label{ft:website}Supplementary video: \href{https://xbpeng.github.io/projects/Robotic_Imitation/}{xbpeng.github.io/projects/Robotic\_Imitation/}}})
\end{abstract}

\IEEEpeerreviewmaketitle

\section{Introduction}
Animals can traverse complex environments with remarkable agility, bringing to bear broad repertoires of agile and acrobatic skills.
Reproducing such agile behaviors has been a long-standing challenge in robotics, with a large body of work devoted to designing control strategies for various \ locomotion \ skills \ \citep{Miura1984DynamicWO,Hopping84,Schwind1998SpringLI,Geyer03,Bledt2018MITC3}. \ However, \\
\\
\vspace{0.1cm}

\vspace{5.47cm}
\noindent designing control strategies often involves a lengthy development process, and requires substantial expertise of both the underlying system and the desired skills. Despite the many success in this domain, the capabilities achieved by these systems are still far from the fluid and graceful motions seen in the animal kingdom.

Learning-based approaches offer the potential to improve the agility of legged robots, while also automating a substantial portion of the manual effort involved in the development of controllers. In particular, reinforcement learning (RL) can be an effective and general approach for developing controllers that can perform a wide range of sophisticated skills \citep{Coros09,2016-TOG-deepRL,HeessTSLMWTEWER17,2018-TOG-deepMimic,BasketballLiu2018}. While these methods have demonstrated promising results in simulation, agents trained through RL are prone to adopting unnatural behaviors that are dangerous or infeasible when deployed in the real world. Furthermore, designing reward functions that elicit the desired behaviors can itself require a laborious task-specific tuning process.

The comparatively superior agility seen in animals, as compared to robots, might lead one to wonder: can we build more agile robotic controllers with less effort by directly \emph{imitating} animal motions? In this work, we propose an imitation learning framework that enables legged robots to learn agile locomotion skills from real-world animals. Our framework leverages reference motion data to provide priors regarding feasible control strategies for a particular skill. The use of reference motions alleviates the need to design skill-specific reward functions, thereby enabling a common framework to learn a diverse array of behaviors. To address the high sample requirements of current RL algorithms, the initial training phase is performed in simulation.
In order to transfer policies learned in simulation to the real world, we propose a sample efficient adaptation technique, which fine-tunes the behavior of a policy

The central contribution of our work is a system that enables legged robots to learn agile locomotion skills by imitating animals. We demonstrate the effectiveness of our framework on a variety of dynamic locomotion skills with the Laikago quadruped robot \citep{laikago}, including different locomotion gaits, as well as dynamic hops and turns.
In our ablation studies, we explore the impact of different design decisions made for the various components of our system.

\section{Related Work}

The development of controllers for legged locomotion has been an enduring subject of interest in robotics, with a large body of work proposing a variety of control strategies for legged systems \citep{Miura1984DynamicWO,Hopping84,Schwind1998SpringLI,Goswami1999,Geyer03,Yin07,2010-TOG-gbwc,Bledt2018MITC3}. However, many of these methods require in-depth knowledge and manual engineering for each behavior, and as such, the resulting capabilities are ultimately limited by the designer's understanding of how to model and represent agile and dynamic behaviors.
Trajectory optimization and model predictive control can mitigate some of the manual effort involved in the design process, but due to the high-dimensional and complex dynamics of legged systems, reduced-order models are often needed to formulate tractable optimization problems \citep{delasa2010,Gehring2016,di2018dynamic,Cassie18}. These simplified abstractions tend to be task-specific, and again require significant insight of the properties of each skill.
\newline
\vspace{-0.2cm}

\noindent \textbf{Motion imitation.}
Imitating reference motions provides a general approach for robots to perform a rich variety of behaviors that would otherwise be difficult to manually encode into controllers \citep{Pollard2002,GrimesCR06,Suleiman08,Yamane2010}. But applications of motion imitation to legged robots have predominantly been limited to behaviors that emphasize upper-body motions, with fairly static lower-body movements, where balance control can be delegated to separate control strategies \citep{Nakaoka2003,KimHumanoid09,Koenemann2014RealtimeIO}. In contrast to physical robots, substantially more dynamic skills can be reproduced by agents in simulation \citep{Muico2009,BipedLee2010,2011-TOG-quadruped,2016-TOG-controlGraphs}. Recently, motion imitation with reinforcement learning has been effective for learning a large repertoire of highly acrobatic skills in simulation \citep{2018-TOG-deepMimic,BasketballLiu2018,2018-TOG-SFV,MuscleConLee2019}. But due to the high sample complexity of RL algorithms and other physical limitations, many of the capabilities demonstrated in simulation have yet to be replicated in the real world.
\newline
\vspace{-0.2cm}

\noindent \textbf{Sim-to-real transfer.}
The challenges of applying RL in the real world have driven the use of domain transfer approaches, where policies are first trained in simulation (source domain), and then transferred to the real world (target domain). Sim-to-real transfer can be facilitated by constructing more accurate simulations \citep{Tan-RSS-18,2019-CORL-cassie}, or adapting the simulator with real-world data \citep{Sim2RealTan2016,AAAI17-Hanna,Hwangboeaau5872,Lowrey18,Chebotar18}. However, building high-fidelity simulators remains a challenging endeavour, and even state-of-the-art simulators provide only a coarse approximation of the rich dynamics of the real world. Domain randomization can be incorporated into the training process to encourage policies to be robust to variations in the dynamics \citep{SadeghiL16,TobinFRSZA17,pinto17a,Sim2Real2018,Andrychowicz2018}. Sample efficient adaptation techniques, such as finetuning \citep{pmlr-v78-rusu17a} and meta-learning \citep{DuanSCBSA16,maml17,clavera2018learning} can also be applied to further improve the performance of pre-trained policies in new domains. In this work, we leverage a class of adaptation techniques, which we broadly referred to as \emph{latent space} methods \citep{He2018,BipedYu19,yu2019learning},
to transfer locomotion policies from simulation to the real world. During pre-training, these methods learn a latent representation of different behaviors that are effective under various scenarios. When transferring to a new domain, a search can be conducted in the latent space to find behaviors that successfully execute a desired task in the new domain. We show that by combining motion imitation and latent space adaptation, our system is able to learn a diverse corpus of dynamic locomotion skills that can be transferred to legged robots in the real world.
\newline
\vspace{-0.2cm}

\noindent \textbf{RL for legged locomotion.}
Reinforcement learning has been effective for automatically acquiring locomotion skills in simulation \citep{2018-TOG-deepMimic,BasketballLiu2018,MuscleConLee2019} and in the real world \citep{KohlS04,Tedrake2004,BipedEndo2005,Tan-RSS-18,Haarnoja18,Hwangboeaau5872}. \citet{KohlS04} applied a policy gradient method to tune manually-crafted walking controllers for the Sony Aibo robot.
By carefully modeling the motor dynamics of the Minitaur quadruped robot, \citet{Tan-RSS-18} was able to train walking policies in simulation that can be directly deployed on a real robot. \citet{Hwangboeaau5872} proposed learning a motor dynamics model using real-world data, which enabled direct transfer of a variety of locomotion skills to the ANYmal robot. Their system trained policies using manually-designed reward functions for each skill, which can be difficult to specify for more complex behaviors. Imitating reference motions can be a general approach for learning diverse repertoires of skills without the need to design skill-specific reward functions \citep{2016-TOG-controlGraphs,2018-TOG-deepMimic,2018-TOG-SFV}.
\citet{2019-CORL-cassie} trained bipedal walking policies for the Cassie robot by imitating reference motions recorded from existing controllers and keyframe animations. The policies are again transferred from simulation to the real world with the aid of careful system identification.
\citet{BipedYu19} transferred bipedal locomotion policies from simulation to a physical Darwin OP2 robot using a latent space adaptation method, which mitigates the dependency on accurate simulators.
In this work, we leverage a similar latent space method, but by combining it with motion imitation, our system enables real robots to perform more diverse and agile behaviors than have been demonstrated by these previous methods.


\section{Overview}
\begin{figure}[t]
	\centering
    \includegraphics[width=1\columnwidth]{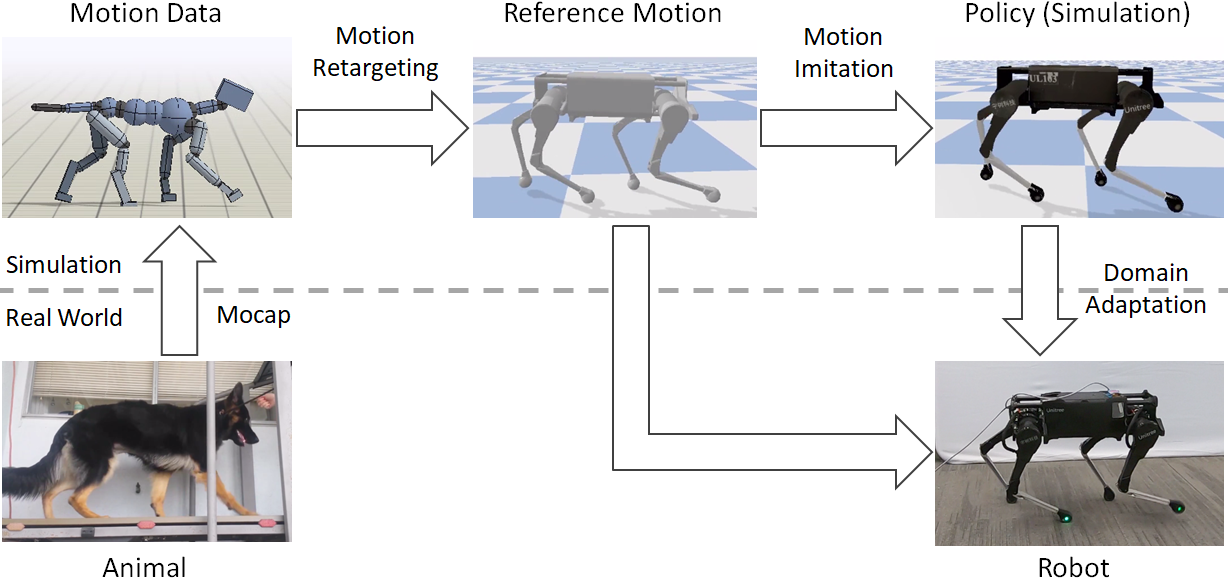}
    \vspace{-0.5cm}
\caption{The framework consists of three stages: motion retargeting, motion imitation, and domain adaptation. It receives as input motion data recorded from an animal, and outputs a control policy that enables a real robot to reproduce the motion. 
}
\label{fig:overview}
\vspace{-0.5cm}
\end{figure}

The objective of our framework is to enable robots to learn skills from real animals. 
Our framework receives as input a reference motion that demonstrates a desired skill for the robot, which may be recorded using motion capture (mocap) of real animals (e.g. a dog). Given a reference motion, it then uses reinforcement learning to synthesize a policy that enables a robot to reproduce that skill in the real world. A schematic illustration of our framework is available in Figure~\ref{fig:overview}. The process is organized into three stages: motion retargeting, motion imitation, and domain adaptation. 1) The reference motion is first processed by the motion retargeting stage, where the motion clip is mapped from the original subject's morphology to the robot's morphology via inverse-kinematics. 2) Next, the retargeted reference motion is used in the motion imitation stage to train a policy to reproduce the motion with a simulated model of the robot. To facilitate transfer to the real world, domain randomization is applied in simulation to train policies that can adapt to different dynamics. 3) Finally, the policy is transferred to a real robot via a sample efficient domain adaptation process, which adapts the policy's behavior using a learned latent dynamics representation.

\section{Motion Retargeting}
\label{sec:retargeting}
\begin{figure}[t]
	\centering
    \subfigure{\includegraphics[width=0.40\columnwidth]{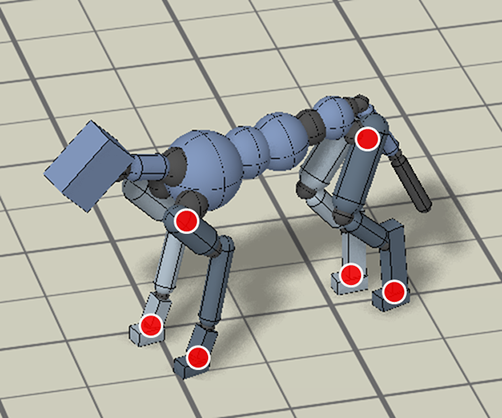}}
    \hspace{0.2cm}
    \subfigure{\includegraphics[width=0.40\columnwidth]{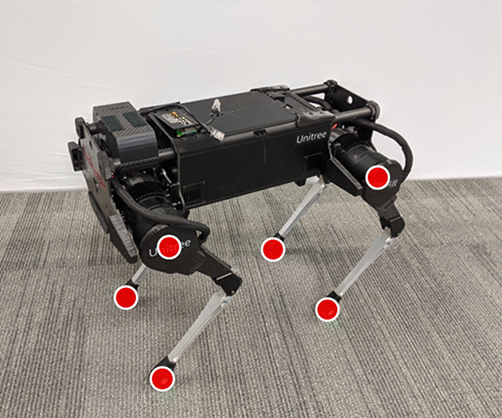}}
\vspace{-0.15cm}
\caption{Inverse-kinematics (IK) is used to retarget mocap clips recorded from a real dog (\textbf{left}) to the Laikago robot (\textbf{right}). Corresponding pairs of keypoints (red) are specified on the dog and robot's bodies, and then IK is used to compute a pose for the robot that tracks the keypoints.}
\label{fig:keypoints}
\vspace{-0.4cm}
\end{figure}

When using motion data recorded from animals, the subject's morphology tends to differ from that of the robot's.
To address this discrepancy, the source motions are retargeted to the robot's morphology using inverse-kinematics \citep{Gleicher1998RMN}.
First, a set of source keypoints are specified on the subject's body, which are paired with corresponding target keypoints on the robot's body. An illustration of the keypoints is available in Figure~\ref{fig:keypoints}. The keypoints include the positions of the feet and hips. At each timestep, the source motion specifies the 3D location $\hat{\rvx}_i(t)$ of each keypoint $i$. The corresponding target keypoint $\rvx_i(\rvq_t)$ is determined by the robot's pose $\rvq_t$, represented in generalized coordinates \citep{Featherstone2007}. IK is then applied to construct a sequence of poses $\rvq_{0:T}$ that track the keypoints at each frame,
\begin{equation}
    \underset{\rvq_{0:T}}{\text{arg min}} \sum_t \sum_i ||\hat{\rvx}_i(t) - \rvx_i(\rvq_t)||^2 + (\bar{\rvq} - \rvq_t)^T \rvW (\bar{\rvq} - \rvq_t).
\end{equation}
An additional regularization term is included to encourage the poses to remain similar to a default pose $\bar{\rvq}$, and ${\rvW = \mathrm{diag}(w_1, w_2, ...)}$ is a diagonal matrix specifying regularization coefficients for each joint.

\section{Motion Imitation}
We formulate motion imitation as a reinforcement learning problem. In reinforcement learning, the objective is to learn a control policy $\pi$ that enables an agent to maximize its expected return for a given task \citep{Sutton1998IRL}. At each timestep $t$, the agent observers a state $\rvs_t$ from the environment, and samples an action $\rva_t \sim \pi(\rva_t| \rvs_t)$ from its policy $\pi$. 
The agent then applies this action, which results in a new state $\rvs_{t+1}$ and a scalar reward $r_t = r(\rvs_t, \rva_t, \rvs_{t+1})$. Repeated applications of this process generates a trajectory $\tau = \{(\rvs_0, \rva_0, r_0), (\rvs_1, \rva_1, r_1), ... \}$. The objective then is to learn a policy that maximizes the agent's expected return $J(\pi)$,
\begin{equation}
    J(\pi) = \expec_{\tau \sim p(\tau | \pi)} \left[\sum_{t=0}^{T-1} \gamma^t r_t \right] ,
\end{equation}
where $T$ denotes the time horizon of each episode, and $\gamma \in [0, 1]$ is a discount factor. $p(\tau | \pi)$ represents the likelihood of a trajectory $\tau$ under a given policy $\pi$,
\begin{equation}
    p(\tau | \pi) = p(\rvs_0) \prod_{t=0}^{T-1} p(\rvs_{t+1} | \rvs_t, \rva_t) \pi(\rva_t | \rvs_t) ,
\end{equation}
with $p(\rvs_0)$ being the initial state distribution, and $p(\rvs_{t+1} | \rvs_t, \rva_t)$ representing the dynamics of the system, which determines the effects of the agent's actions.

To imitate a given reference motion, we follow a similar motion imitation approach as \citet{2018-TOG-deepMimic}. The inputs to the policy is augmented with an additional goal $\rvg_t$, which specifies the motion that the robot should imitate. The policy is modeled as a feedforward network that maps a given state $\rvs_t$ and goal $\rvg_t$ to a distribution over actions $\pi(\rva_t | \rvs_t, \rvg_t)$. The policy is queried at 30Hz for a new action at each timestep. The state ${\rvs_t = (\rvq_{t-2:t}, \rva_{t-3:t-1})}$ is represented by the poses $\rvq_{t-2:t}$ of the robot in the three previous timesteps, and the three previous actions $\rva_{t-3:t-1}$. The pose features $\rvq_t$ consist of IMU readings of the root orientation (row, pitch, yaw) and the local rotations of every joint. The root position is not included among the pose features to avoid the need to estimate the root position during real-world deployment. The goal $\rvg_t = (\hat{\rvq}_{t+1}, \hat{\rvq}_{t+2}, \hat{\rvq}_{t+10}, \hat{\rvq}_{t+30})$ specifies target poses from the reference motion at four future timesteps, spanning approximately 1 second. The action $\rva_t$ specifies target rotations for PD controllers at each joint. To ensure smoother motions, the PD targets are first processed by a low-pass filter before being applied on the robot \citep{butterworth1930theory}.
\newline

\noindent \textbf{Reward Function.}
The reward function encourages the policy to track the sequence of target poses $(\hat{\rvq}_0, \hat{\rvq}_1, ..., \hat{\rvq}_T)$ from the reference motion at every timestep. The reward function is similar to the one used by \citet{2018-TOG-deepMimic}, where the reward $r_t$ at each timestep is given by:
\begin{equation}
    r_t = w^\text{p} r^\text{p}_t + w^\text{v} r^\text{v}_t + w^\text{e} r^\text{e}_t + w^\text{rp} r^\text{rp}_t  + w^\text{rv} r^\text{rv}_t
\end{equation}
\[
    w^\text{p}=0.5, \ w^\text{v}=0.05, \ w^\text{e}=0.2, \ w^\text{rp}=0.15, \ w^\text{rv}=0.1
\]
The pose reward $r^\text{p}_t$ encourages the robot to minimize the difference between the joint rotations specified by the reference motion and those of the robot. In the equation below, $\hat{\rvq}_t^j$ represents the 1D local rotation of joint $j$ from the reference motion at time $t$, and $\rvq_t^j$ represents the robot's joint,
\begin{equation}
    r^\text{p}_t = \mathrm{exp}\left[ -5 \sum_j ||\hat{\rvq}_t^j - \rvq_t^j ||^2\right].
\end{equation}
Similarly, the velocity reward $r^\text{v}_t$ is calculated according to the joint velocities, with $\hat{\dot{\rvq}}_t^j$ and $\dot{\rvq}_t^j$ being the angular velocity of joint $j$ from the reference motion and robot respectively,
\begin{equation}
    r^\text{v}_t = \mathrm{exp}\left[ -0.1 \sum_j ||\hat{\dot{\rvq}}_t^j - \dot{\rvq}_t^j ||^2\right].
\end{equation}
Next, the end-effector reward $r^\text{e}_t$, encourages the robot to track the positions of the end-effectors, where $\rvx_t^e$ denotes the relative 3D position of end-effector $e$ with respect to the root,
\begin{equation}
    r^\text{e}_t = \mathrm{exp}\left[ -40 \sum_e ||\hat{\rvx}_t^e - \rvx_t^e ||^2\right].
\end{equation}
Finally, the root pose reward $r^\text{rp}_t$ and root velocity reward $r^\text{rv}_t$ encourage the robot to track the reference root motion. $\rvx_t^\text{root}$ and $\dot{\rvx}_t^\text{root}$ denotes the root's global position and linear velocity, while $\rvq_t^\text{root}$ and $\dot{\rvq}_t^\text{root}$ are the rotation and angular velocity,
\begin{align}
    r^\text{rp}_t = \mathrm{exp}\left[-20||\hat{\rvx}_t^\text{root} - \rvx_t^\text{root}||^2 - 10 ||\hat{\rvq}_t^\text{root} - \rvq_t^\text{root}||^2 \right] \ \\
    r^\text{rv}_t = \mathrm{exp}\left[-2 ||\hat{\dot{\rvx}}_t^\text{root} - \dot{\rvx}_t^\text{root}||^2 - 0.2||\hat{\dot{\rvq}}_t^\text{root} - \dot{\rvq}_t^\text{root}||^2 \right] .
\end{align}

\section{Domain Adaptation}

Due to discrepancies between the dynamics of the simulation and the real world, policies trained in simulation tend to perform poorly when deployed on a physical system. Therefore, we propose a sample efficient adaptation technique for transferring policies from simulation to the real world.

\subsection{Domain Randomization}
Domain randomization is a simple strategy for improving a policy's robustness to dynamics variations \citep{SadeghiL16,TobinFRSZA17,Sim2Real2018}. Instead of training a policy in a single environment with fixed dynamics, domain randomization varies the dynamics during training, thereby encouraging the policy to learn strategies that are functional across different dynamics. However, there may be no single strategy that is effective across all environments, and due to unmodeled effects in the real world, strategies that are robust to different simulated dynamics may nonetheless fail when deployed in a physical system. 

\subsection{Domain Adaptation} 
In this work, we aim to learn strategies that are robust to variations in the dynamics of the environment, while also being able to adapt its behaviors as necessary for new environments.
Let $\rvmu$ represent the values of the dynamics parameters that are randomized during training in simulation (Table \ref{tab:dynamicsParams}). At the start of each episode, a random set of parameters are sampled according to $\rvmu \sim p(\rvmu)$. The dynamics parameters are then encoded into a latent embedding $\rvz \sim E(\rvz | \rvmu)$ by a stochastic encoder $E$,
and $\rvz$ is provided as an additional input to the policy $\pi(\rva | \rvs, \rvz)$. For brevity, we have excluded the goal input $\rvg$ for the policy.
When transferring a policy to the real world, we follow a similar approach as \citet{yu2018policy}, where a search is performed to find a latent encoding $\rvz^*$ that enables the policy to successfully execute the desired behaviors on the physical system. Next, we propose an extension that addresses potential issues due to over-fitting with the previously proposed method.

A potential degeneracies of the previously described approach is that the policy may learn strategies that depend on $\rvz$ being an accurate representation of the true dynamics of the system. This can result in brittle behaviors where the strategies utilized by the policy for a given $\rvz$ can overfit to the precise dynamics from the corresponding parameters $\rvmu$. Furthermore, due to unmodeled effects in the real world, there might be no $\rvmu$ that accurately models real-world dynamics. Therefore, to encourage the policy to be robust to uncertainty in the dynamics, we incorporate an information bottleneck into the encoder. The information bottleneck enforces an upper bound $I_c$ on the mutual information $I(\rvM, \rvZ)$ between the dynamics parameters $\rvM$ and the encoding $\rvZ$.
This results in the following constrained policy optimization objective,
\begin{align}
& \underset{\pi, E}{\text{arg max}}
& & \expec_{\rvmu \sim p(\rvmu)} \expec_{\rvz \sim E(\rvz | \rvmu)} \expec_{\tau \sim p(\tau | \pi, \rvmu, \rvz)} \left[\sum_{t=0}^{T-1} \gamma^t r_t \right] \\
& \text{s.t.}
& & I(\rvM, \rvZ) \leq I_c . \label{eqn:VIBCons}
\end{align}
where the trajectory distribution is now given by,
\begin{equation}
    p(\tau | \pi, \rvmu, \rvz) = p(\rvs_0) \prod_{t=0}^{T-1} p(\rvs_{t+1} | \rvs_t, \rva_t, \rvmu) \pi(\rva_t | \rvs_t, \rvz) .
\end{equation}
Since computing the mutual information is intractable, the constraint in Equation~\ref{eqn:VIBCons} can be approximated with a variational upper bound using the KL divergence between $E$ and a variational prior $\rho(\rvz)$ \citep{AlemiFD016},
\begin{equation}
    I(\rvM, \rvZ) \leq \expec_{\rvmu \sim p(\rvmu)} \left[\mathrm{D}_\mathrm{KL} \left[E(\cdot | \rvmu) || \rho(\cdot)\right]\right] .
\end{equation}
We can further simplify the objective by converting Equation~\ref{eqn:VIBCons} into a soft constraint, to yield the following information-regularized objective,
\begin{equation}
\begin{aligned}
\underset{\pi, E}{\text{arg max}} \
& \expec_{\rvmu \sim p(\rvmu)} \expec_{\rvz \sim E(\rvz | \rvmu)} \expec_{\tau \sim p(\tau | \pi, \rvmu, \rvz)} \left[\sum_{t=0}^{T-1} \gamma^t r_t \right] \\
& - \beta \ \expec_{\rvmu \sim p(\rvmu)} \left[\mathrm{D}_\mathrm{KL} \left[E(\cdot | \rvmu) || \rho(\cdot)\right] \right] ,
\end{aligned}
\end{equation}
with $\beta \geq 0$ being a Lagrange multiplier. In our experiments, we model the encoder $E(\rvz | \rvmu) = \mathcal{N}\left(\rvm(\rvmu), \rvSigma(\rvmu)\right)$ as a Gaussian distribution with mean $\rvm(\rvmu)$ and standard deviation $\rvSigma(\rvmu)$, and the prior $\rho(\rvz) = \mathcal{N}\left(0, \rvI\right)$ is given by the unit Gaussian. This objective can be interpreted as training a policy that maximizes the agent's expected return across different dynamics, while also being able to adapt its behaviors when necessary by relying on only a minimal amount of information from the ground-truth dynamics parameters. 
In our formulation, the Lagrange multiplier $\beta$ provides a trade-off between robustness and adaptability. Large values of $\beta$ restrict the amount of information that the policy can access from $\rvmu$. In the limit $\beta \rightarrow \infty$, the policy converges to a robust but non-adaptive policy that does not access the underlying dynamics parameters. Conversely, small values of $\beta \rightarrow 0$ provides the policy with unfettered access to the dynamics parameters, which can result in brittle strategies where the policy's behaviors overfit to the nuances of each setting of the dynamics parameters, potentially leading to poor generalization to real-world dynamics. 

\begin{algorithm}[t]
            \caption{Adaptation with Advantage-Weighted Regression}
            \begin{algorithmic}[1]
            
            \STATE{$\pi \leftarrow \text{trained policy}$}
            \STATE{$\omega_0 \leftarrow \mathcal{N}(0, I)$}
            \STATE{$\mathcal{D} \leftarrow \emptyset$}
            
            \FOR{$\text{iteration} \ k = 0, ... , k_\mathrm{max} - 1$}
                \STATE{$\rvz_k \leftarrow \text{sampled encoding from} \ \omega_k(\rvz)$}
                \STATE{Rollout an episode with $\pi$ conditioned $\rvz_k$ and record the return $\mathcal{R}_k$}
                \STATE{Store $(\rvz_k, \mathcal{R}_k)$ in $\mathcal{D}$}
                
                \STATE{$\bar{v} \leftarrow \frac{1}{k} \sum_{i=1}^k \mathcal{R}_i$}
                \STATE{$\omega_{k+1} \leftarrow \mathop{\mathrm{arg\ max}}_{\omega} \sum_{i=1}^k \left[ \mathrm{log} \ \omega(\rvz_i) \ \mathrm{exp}\left( \frac{1}{\alpha} \left(\mathcal{R}_i - \bar{v} \right) \right) \right]$} \label{alg:adaptAWR_update}
                \vspace{-0.3cm}
            \ENDFOR
            \end{algorithmic}
            \label{alg:adaptAWR}
\end{algorithm}

\subsection{Real World Transfer}
To adapt a policy to the real world, we directly search for an encoding $\rvz$ that maximizes the return on the physical system
\begin{equation}
\rvz^* = \underset{\rvz}{\text{arg max}} \quad \expec_{\tau \sim p^*(\tau | \pi, \rvz)} \left[\sum_{t=0}^{T-1} \gamma^t r_t \right] ,
\end{equation}
with $p^*(\tau | \pi, \rvz)$ being the trajectory distribution under real-world dynamics. To identify $\rvz^*$, we use advantage-weighted regression (AWR) \citep{FQIbyAWR2009,AWRPeng19}, a simple off-policy RL algorithm.
Algorithm~\ref{alg:adaptAWR} summarizes the adaptation process. The search distribution is initialized with the prior $\omega_0(\rvz) = \mathcal{N}(0, I)$. At each iteration $k$, we sample an encoding from the current distribution $\rvz_k \sim \omega_k(\rvz)$ and execute
an episode with the policy conditioned on $\rvz_k$. The return $\mathcal{R}_k$ for the episode is recorded and stored along with $\rvz_k$ in a replay buffer $\mathcal{D}$ containing all samples from previous iterations. $\omega_k(\rvz)$ is then updated by fitting a new distribution that assigns higher likelihoods to samples with larger advantages. The likelihood of each sample $\rvz_i$ is weighted by the exponentiated-advantage $\mathrm{exp}\left(\frac{1}{\alpha}\left(\mathcal{R}_i - \bar{v} \right)\right)$, where the baselines $\bar{v}$ is the average return of all samples in $\mathcal{D}$, and $\alpha$ is a manually specified temperature parameter. Note that, since $\omega_k(\rvz)$ is Gaussian, the optimal distribution at each iteration (Line~\ref{alg:adaptAWR_update}) can be determined analytically. However, we found that the analytic solution is prone to premature convergence to a suboptimal solution. Instead, we update $\omega_k(\rvz)$ incrementally using a few steps of gradient descent. This process is repeated for $k_\mathrm{max}$ iterations, and the mean of the final distribution $\omega_{k_\mathrm{max}}(\rvz)$ is used as an approximation of the optimal encoding $\rvz^*$ for deploying the policy in the real world.

\begin{table}[t]
\centering 
\resizebox{\columnwidth}{!}{
\begin{tabular}{|l|c|c|}
\hline
{\bf Parameter} & {\bf Training Range} & {\bf Testing Range} \\ \hline
Mass & $[0.8, 1.2]\ \times$ default value & $[0.5, 2.0]\ \times$ default value \\ \hline
Inertia & $[0.5, 1.5]\ \times$ default value & $[0.4, 1.6]\ \times$ default value \\ \hline
Motor Strength & $[0.8, 1.2]\ \times$ default value & $[0.7, 1.3]\ \times$ default value \\ \hline
Motor Friction & $[0, 0.05]\ N m s / rad$ & $[0, 0.075]\ N m s / rad$ \\ \hline
Latency & $[0, 0.04]\ s$ &  $[0, 0.05]\ s$ \\ \hline
Lateral Friction & $[0.05, 1.25]\ N s / m$ & $[0.04, 1.35]\ N s / m$ \\ \hline
\end{tabular}
}
\caption{Dynamic parameters and their respective range of values used during training and testing. A larger range of values are used during testing to evaluate the policies' ability to generalize to unfamiliar dynamics.}
\label{tab:dynamicsParams}
\vspace{-0.8cm}
\end{table}

\begin{figure*}[t!]
    \centering
    \subfigure[Dog Pace \label{fig:snapshotsPace}]{\includegraphics[width=0.49\textwidth]{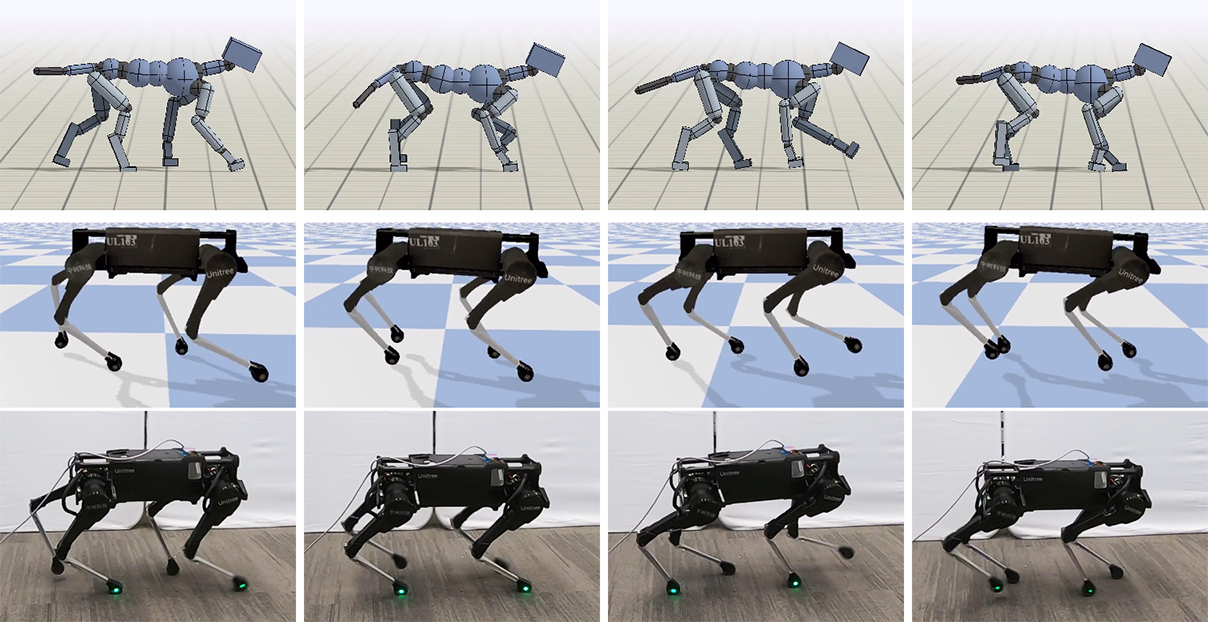}}
    \hfill
    \subfigure[Dog Backwards Trot \label{fig:snapshotsTrotBackward}]{\includegraphics[width=0.49\textwidth]{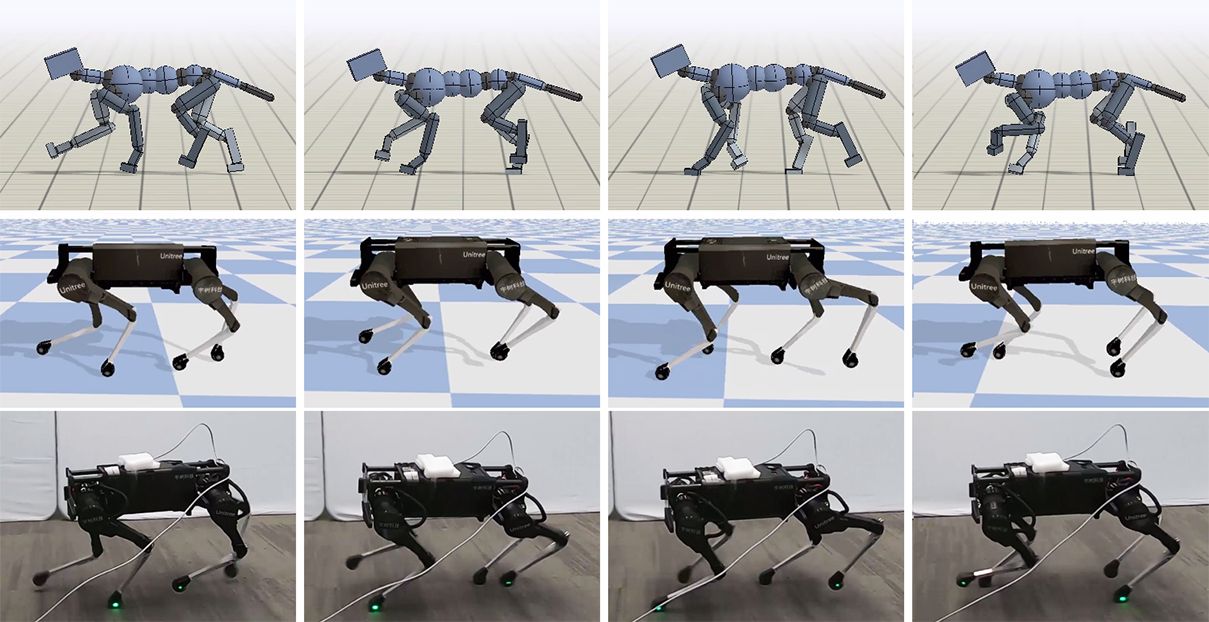}} \\
    \vspace{-0.3cm}
    \subfigure[Side-Steps]{\includegraphics[width=0.49\textwidth]{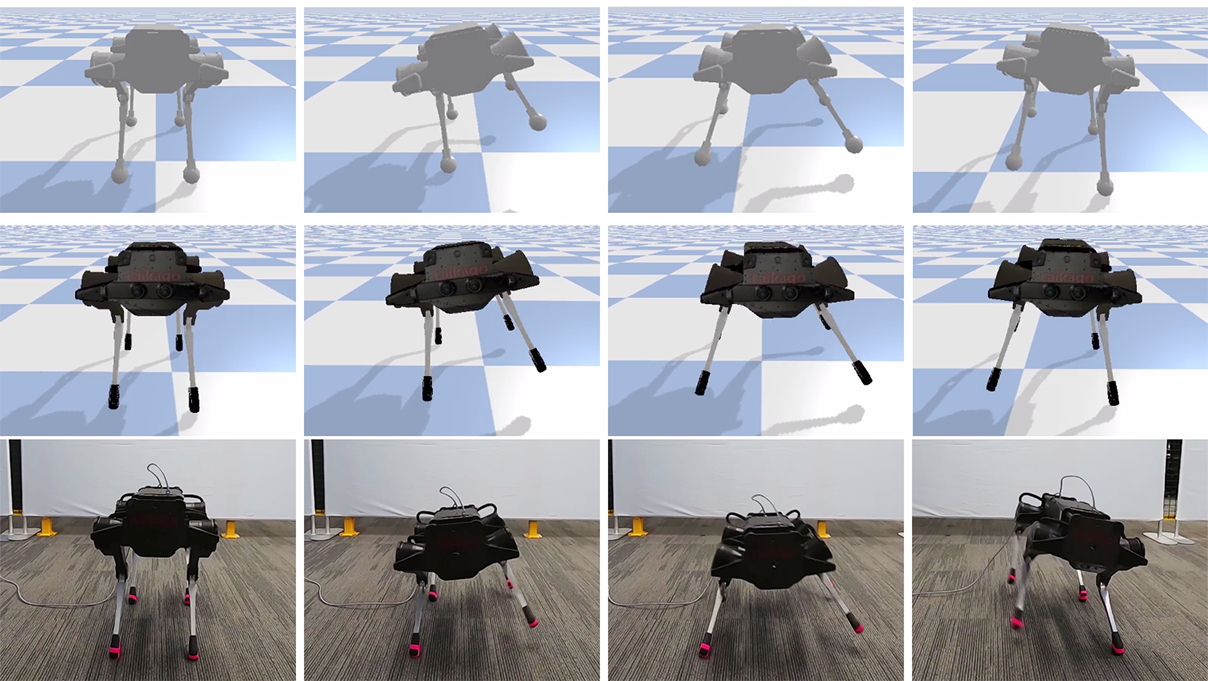}}
    \hfill
    \subfigure[Turn]{\includegraphics[width=0.49\textwidth]{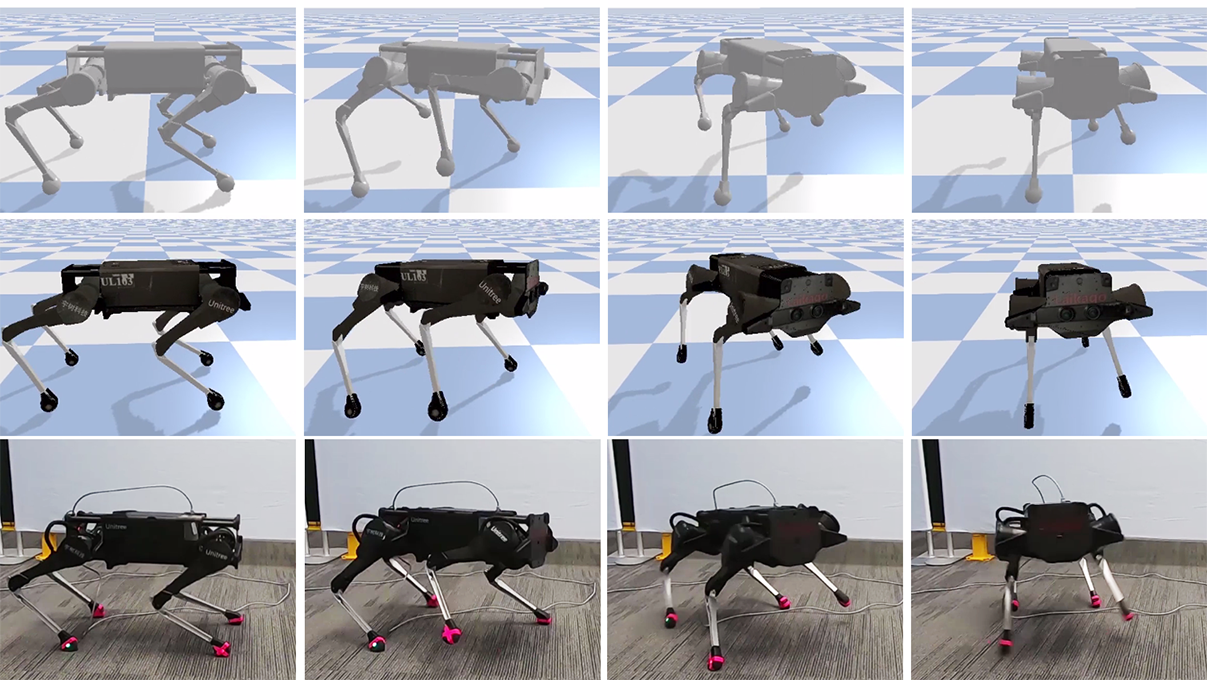}} \\
    \vspace{-0.3cm}
    \subfigure[Hop-Turn \label{fig:snapshotsHopTurn}]{\includegraphics[width=0.49\textwidth]{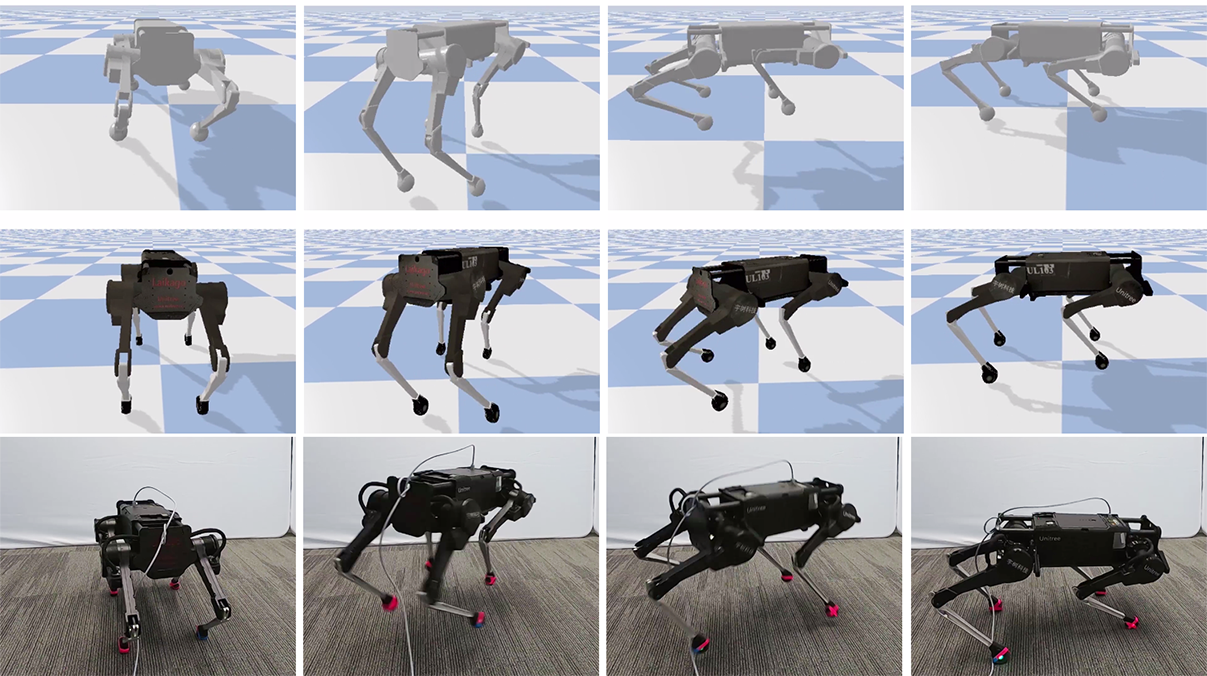}}
    \hfill
    \subfigure[Running Man \label{fig:snapshotsRunningMan}]{\includegraphics[width=0.49\textwidth]{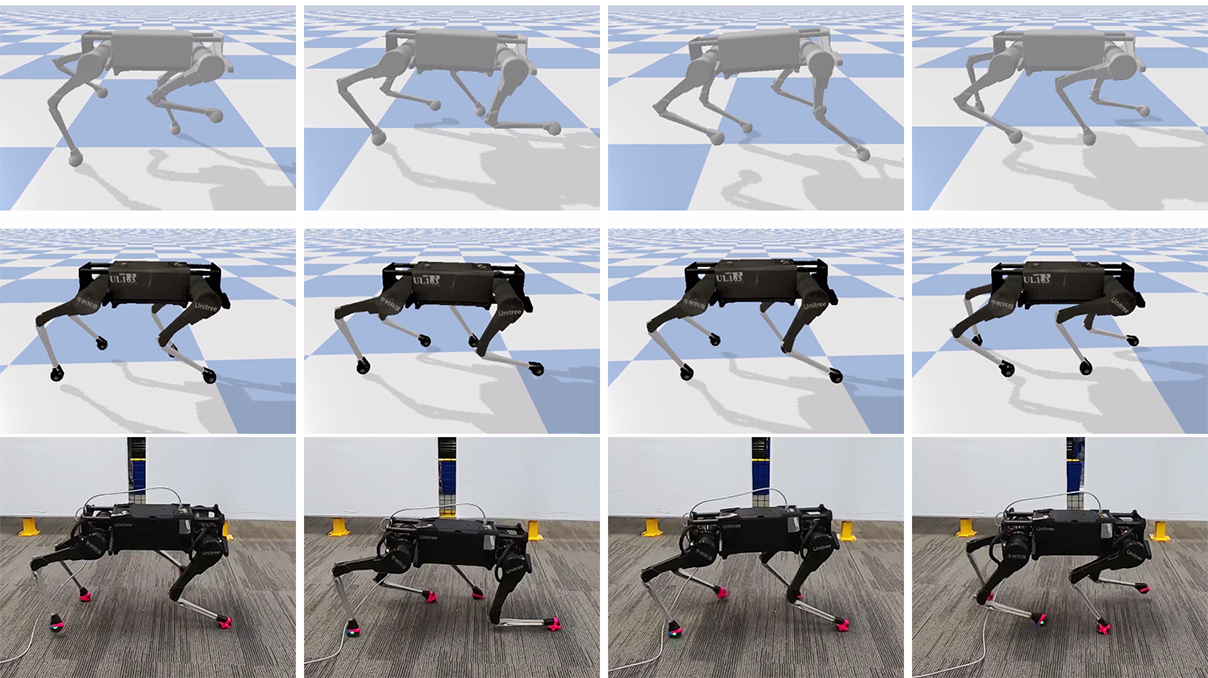}} \\
    \vspace{-0.2cm}
\caption{Laikago robot performing skills learned by imitating reference motions. \textbf{Top:} Reference motion. \textbf{Middle:} Simulated robot. \textbf{Bottom:} Real robot.}
\label{fig:snapshots}
\vspace{-0.5cm}
\end{figure*}

\section{Experimental Evaluation}

We evaluate our robotic learning system by learning to imitating a variety of dynamic locomotion skills using the Laikago robot \citep{laikago}, an 18 degrees-of-freedom quadruped with 3 actuated degrees-of-freedom per leg, and 6 under-actuated degrees of freedom for the root (torso). Behaviors learned by the policies are best seen in the \href{https://xbpeng.github.io/projects/Robotic_Imitation/}{supplementary video\footref{ft:website}}, and snapshots of the behaviors are also available in Figure~\ref{fig:snapshots}. In the following experiments, we aim to evaluate the effectiveness of our framework on learning a diverse set of quadruped skills, and study how well real-world adaptation can enable more agile behaviors. We show that our adaptation method can efficiently transfer policies trained in simulation to the real world with a small number of trials on the physical system. We further study the effects of regularizing the latent dynamics encoding with an information bottleneck, and show that this provides a mechanism to trade off between the robustness and adaptability of the learned policies.

\subsection{Experimental Setup}

\begin{figure*}[t!]
    \centering
    \includegraphics[width=0.9\textwidth]{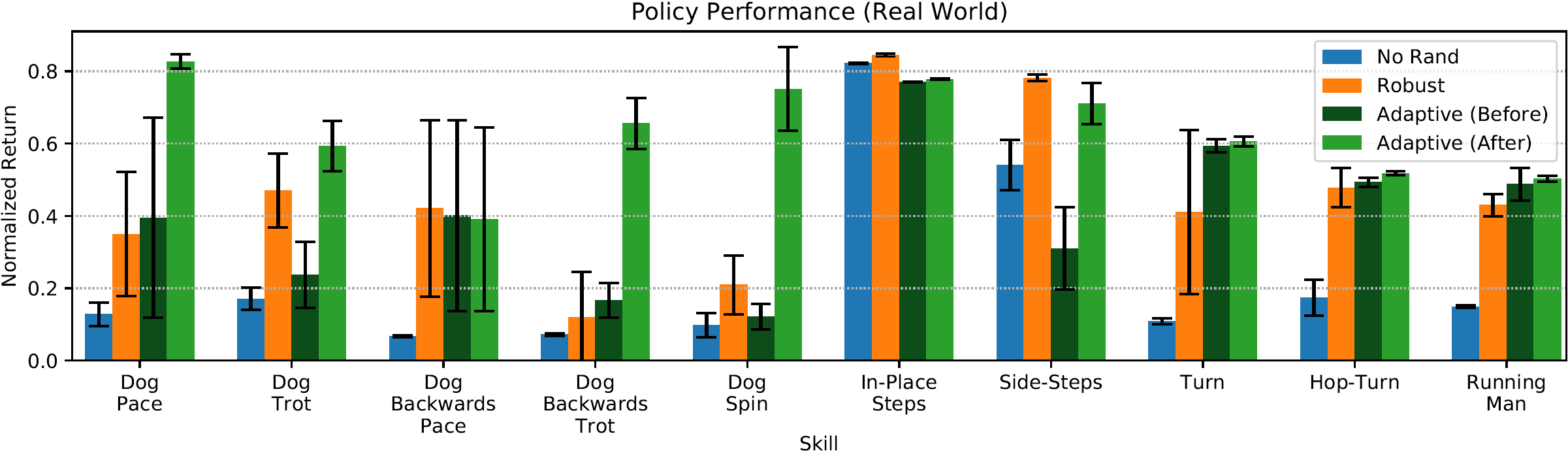}
    \vspace{-0.3cm}
\caption{Performance statistics of imitating various skills in the real world. Performance is recorded as the average normalized return between [0, 1]. Three policies initialized with different random seeds are trained for each combination of skill and method. The performance of each policy is evaluated over 5 episodes, for a total of 15 trials per method. The adaptive policies outperform the non-adaptive policies on most skills.}
\label{fig:perf}
\vspace{-0.35cm}
\end{figure*}

Retargeting via inverse-kinematics and simulated training is performed using PyBullet \citep{coumans2019}. Table~\ref{tab:dynamicsParams} summarizes the dynamics parameters and their respective range of values. The motion dataset contains a mixture of mocap clips recorded from a dog and clips from artist generated animations. The mocap clips are collected from a public dataset \citep{Zhang2018MNN} and retargeted to the Laikago following the procedure in Section~\ref{sec:retargeting}. Figure~\ref{fig:perf} lists the skills learned by the robot and summarizes the performance of the policies when deployed in the real world. Motion clips recorded from a dog are designated with ``Dog", and the other clips correspond to artist animated motions.
Performance is recorded as the average normalized return, with 0 corresponding to the minimum possible return per episode and 1 being the maximum return. Note that the maximum return may not be achievable, since the reference motions are generally not physically feasible for the robot. Performance is calculated using the average of 3 policies initialized with different random seeds. Each policy is trained with proximal policy optimization using about 200 million samples in simulation \citep{PPOSchulmanWDRK17}. Both the encoder and policy are trained end-to-end using the reparameterization trick \citep{KingmaW13}. Domain adaptation is performed on the physical system with AWR in the latent dynamics space, using approximately 50 real-world trials to adapt each policy. Trials vary between 5s and 10s in length depending on the space requirements of each skill. Hyperparameter settings are available in Appendix~\ref{app:hyperParams}.
\newline
\vspace{-0.3cm}

\begin{figure}[t]
    \centering
    \includegraphics[width=0.95\columnwidth]{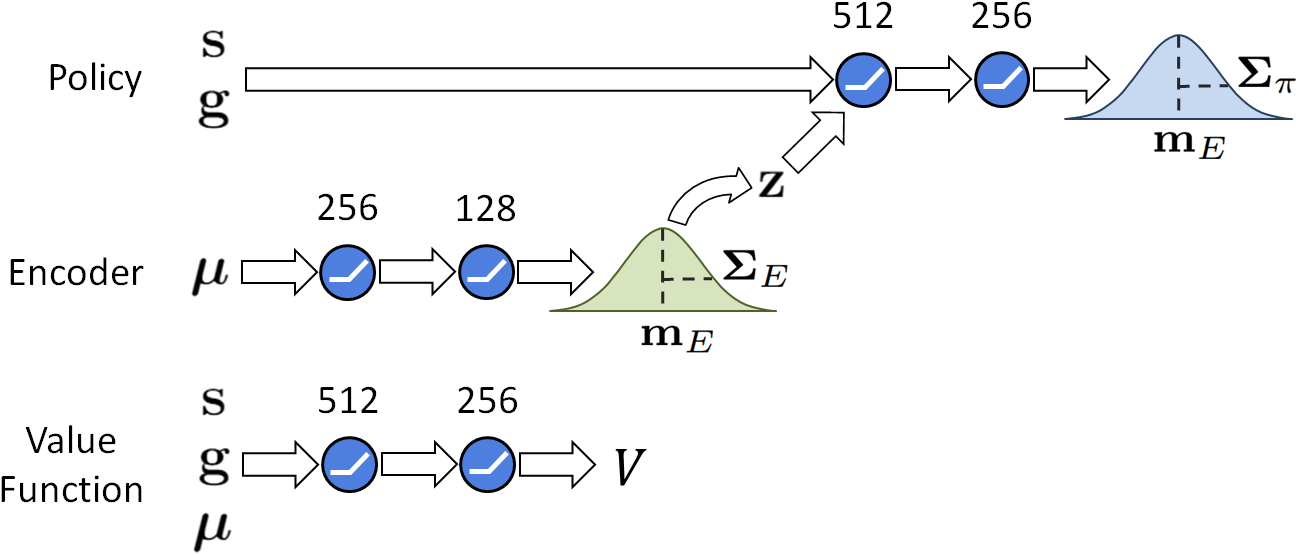}
    \vspace{-0.1cm}
    \caption{Schematic illustration of the network architecture used for the adaptive policy. The encoder $E(\rvz | \rvmu)$ receives the dynamics parameters $\rvmu$ as input, which are processed by two fully-connected layers with 256 and 128 ReLU units, and then mapped to a Gaussian distribution over the latent space $\rvZ$ with mean $\rvm_E(\rvmu)$ and standard deviation $\rvSigma_E(\rvmu)$. An encoding $\rvz$ is sampled from the encoder distribution and provided to the policy $\pi(\rva | \rvs, \rvg, \rvmu)$ as input, along with the state $\rvs$ and goal $\rvg$. The policy is modeled with two layers of 512 and 256 units, followed by an output layer which specifies the mean $\rvm_\pi(\rvs, \rvg, \rvz)$ of the action distribution. The standard deviation $\rvSigma_\pi$ of the action distribution is specified by a fixed diagonal matrix. The value function $V(\rvs, \rvg, \rvmu)$ is modeled by a separate network with 512 and 256 hidden units. 
    }
    \label{fig:net}
    \vspace{-0.5cm}
\end{figure}

\noindent \textbf{Model representation.}
All policies are modeled using the neural network architecture shown in Figure~\ref{fig:net}. The encoder $E(\rvz | \rvmu)$ is represented by a fully-connected network that maps the dynamics parameters $\rvmu$ to the mean $\rvm_E(\rvmu)$ and standard deviation $\rvSigma_E(\rvmu)$ of the encoder distribution. The policy network $\pi(\rva | \rvs, \rvg, \rvz)$ receives as input the state $\rvs$, goal $\rvg$, and dynamics encoding $\rvz$, then outputs the mean $\rvm_\pi(\rvs, \rvg, \rvz)$ of a Gaussian action distribution. The standard deviation $\rvSigma_\pi = \mathrm{diag}(\sigma_\pi^1, \sigma_\pi^2, ...)$ of the action distribution is represented by a fixed matrix. The value function $V(\rvs, \rvg, \rvmu)$ receives as input the state, goal, and dynamics parameters.

\subsection{Learned Skills}
Our framework is able to learn a diverse set of locomotion skills for the Laikago, including dynamic gaits, such as pacing and trotting, as well as agile turning and spinning motions (Figure~\ref{fig:snapshots}).
Pacing is typically used for walking at slower speeds, and is characterized by each pair of legs on the same side of the body moving in unison (Figure~\ref{fig:snapshotsPace}) \citep{raibert1990trotting}. Trotting is a faster gait, where diagonal pairs of legs move together (Figure~\ref{fig:teaser}). We are able to train policies for these different gaits just by providing the system with different reference motions. Furthermore, by simply playing the mocap clips backwards, we are able to train policies for different backwards walking gaits (Figure~\ref{fig:snapshotsTrotBackward}). The gaits learned by our policies are faster than those of the manually-designed controller from the manufacturer. The fastest manufacturer gait reaches a top speed of about 0.84m/s, while the Dog Trot policy reaches a speed of 1.08m/s. The backwards trotting gait reaches an even higher speed of 1.20m/s. In addition to imitating mocap data from animals, our system is also able to learn from artist animated motions. While these hand-animated motions are generally not physically correct, the policies are nonetheless able to closely imitate most motions with the real robot. This includes a highly dynamic Hop-Turn motion,
in which the robot performs a 90 degrees turn midair (Figure~\ref{fig:snapshotsHopTurn}). While our system is able to imitate a variety of motions, some motions, such as Running Man (Figure~\ref{fig:snapshotsRunningMan}), prove challenging to reproduce.
The motion requires the robot to travel backwards while moving in a forward-walking manner. Our policies learn to keep the robot's feet on the ground and shuffle backwards, instead of lifting the feet during each step. 

\begin{figure}[t]
    \centering
    \includegraphics[width=0.9\columnwidth]{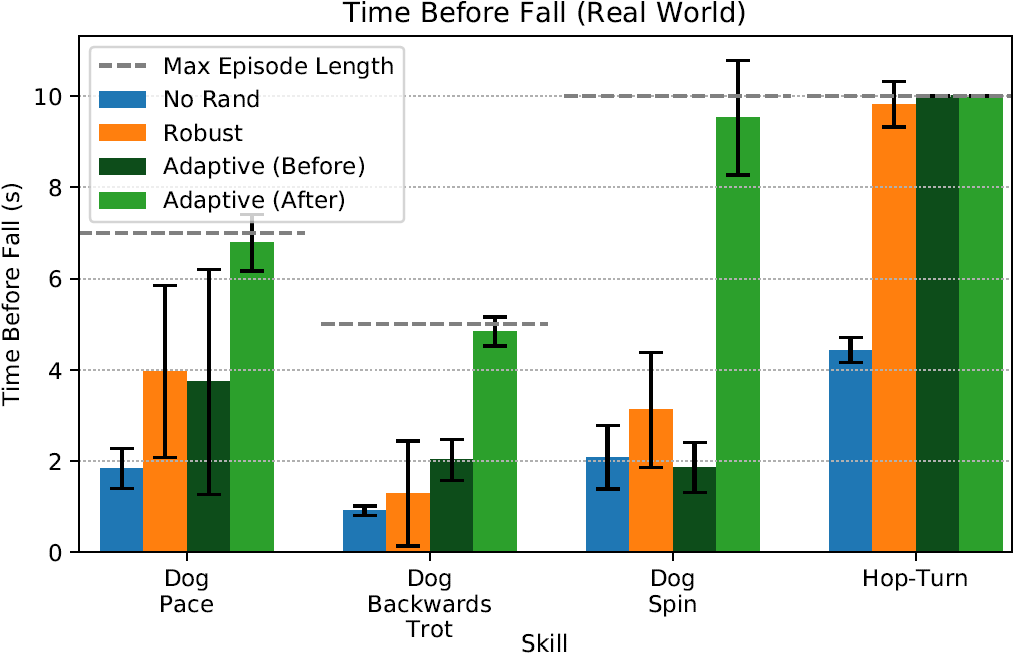}
    \vspace{-0.25cm}
\caption{Comparison of the time elapsed before the robot falls when deploying various policies in the real world. The adaptive policies are often able to maintain balance longer than the other baselines policies, and tend to reach the max episode length without falling.}
\label{fig:fallTime}
\vspace{-0.7cm}
\end{figure}

\subsection{Domain Adaptation}
To determine the effects of domain adaptation, we compare our method to non-adaptive policies trained in simulation without randomization (No Rand), and robust policies trained with randomization (Robust) but do not perform adaptation in new environments. Real-world performance comparisons of these methods are shown in Figure~\ref{fig:perf}, detailed performance statistics in simulation and the real world are available in Appendix~\ref{app:perfStats}. When deployed on the real robot, the adaptive policies outperform their non-adaptive counterparts on most skills. For simpler skills, such as In-Place Steps and Side-Steps, the robust policies are sufficient for transfer to the real robot. But for more dynamic skills, such as Dog Pace and Dog Spin, the robust policies are prone to falling, while the adaptive policies can execute the skills more consistently. Policies trained without randomization fail to transfer to the real world for most skills. Figure~\ref{fig:fallTime} compares the time elapsed before the robot falls under the various policies. The adaptive policies are often able to maintain balance for a longer period of time than the other methods, with a significant performance improvement after adaptation.

\begin{figure}[t]
	\centering
    \subfigure{\includegraphics[height=0.35\columnwidth]{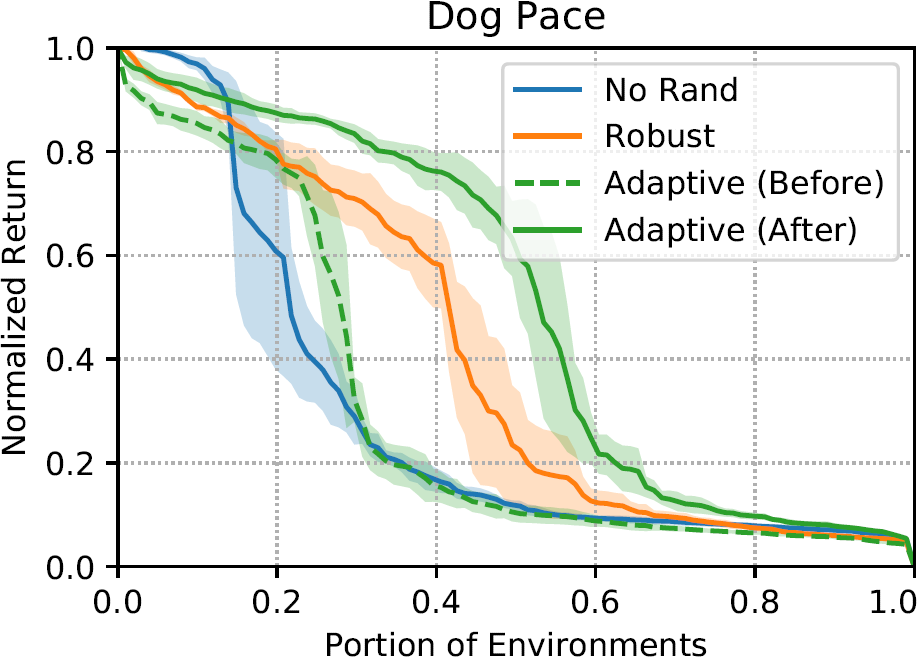}}
    \hfill
    \subfigure{\includegraphics[height=0.35\columnwidth]{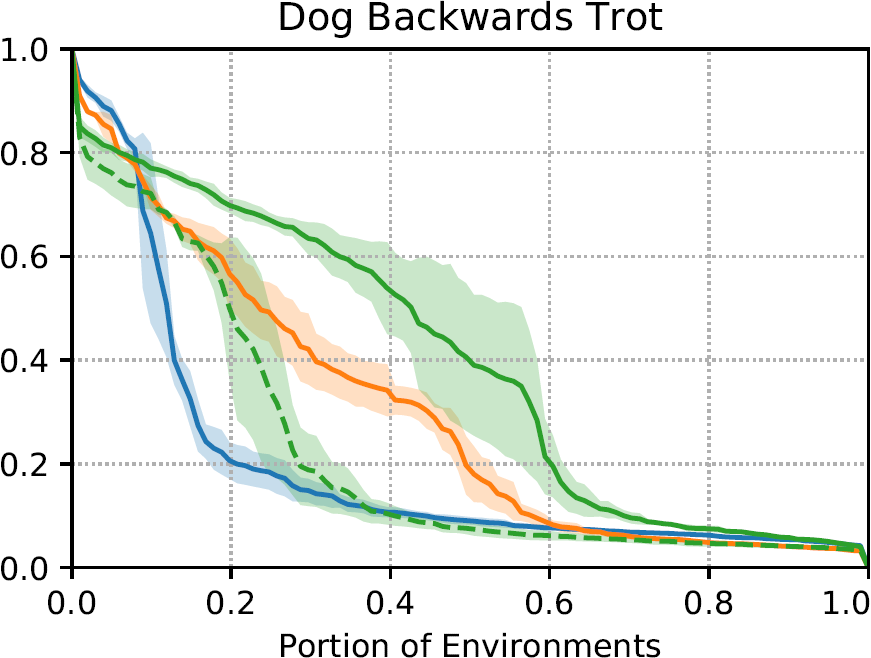}}\\\vspace{-0.15cm}
    \subfigure{\includegraphics[height=0.35\columnwidth]{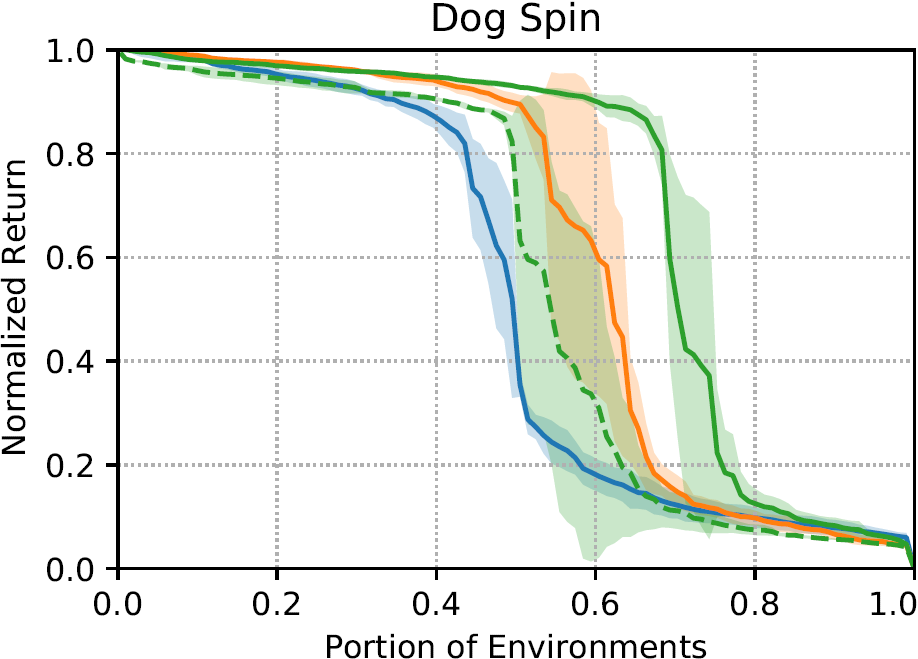}}
    \hfill
    \subfigure{\includegraphics[height=0.35\columnwidth]{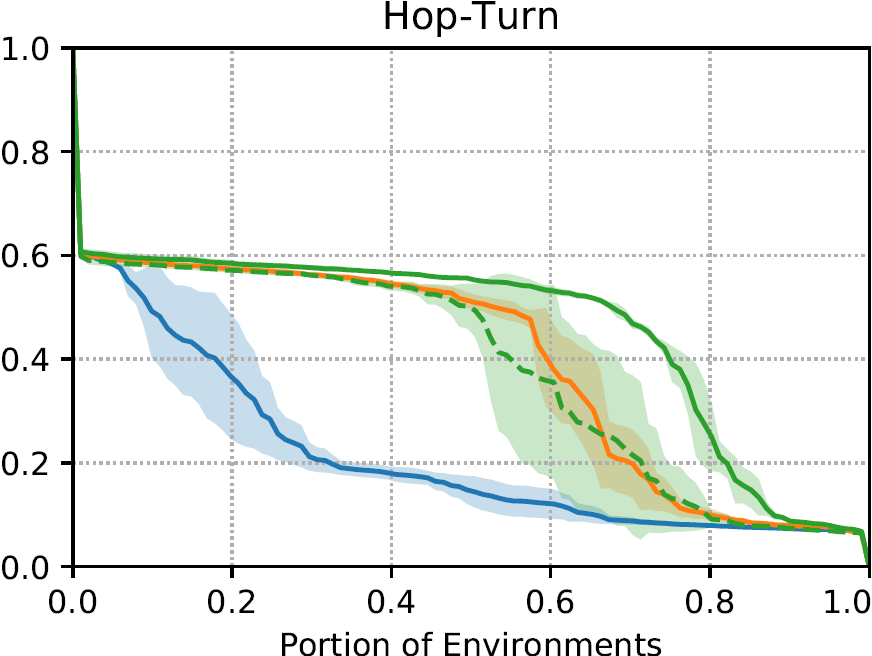}}
    \vspace{-0.2cm}
\caption{Performance of policies in 100 simulated environments with different dynamics. The vertical axis represents the normalized return, and the horizontal axis records the portion of environments in which a policy achieves a return higher than a particular value.
The adaptive policies achieve higher returns under more diverse dynamics than the non-adaptive policies.}
\label{fig:adaptationCurves}
\vspace{-0.4cm}
\end{figure}

\begin{figure}[t]
	\centering
    \subfigure{\includegraphics[height=0.35\columnwidth]{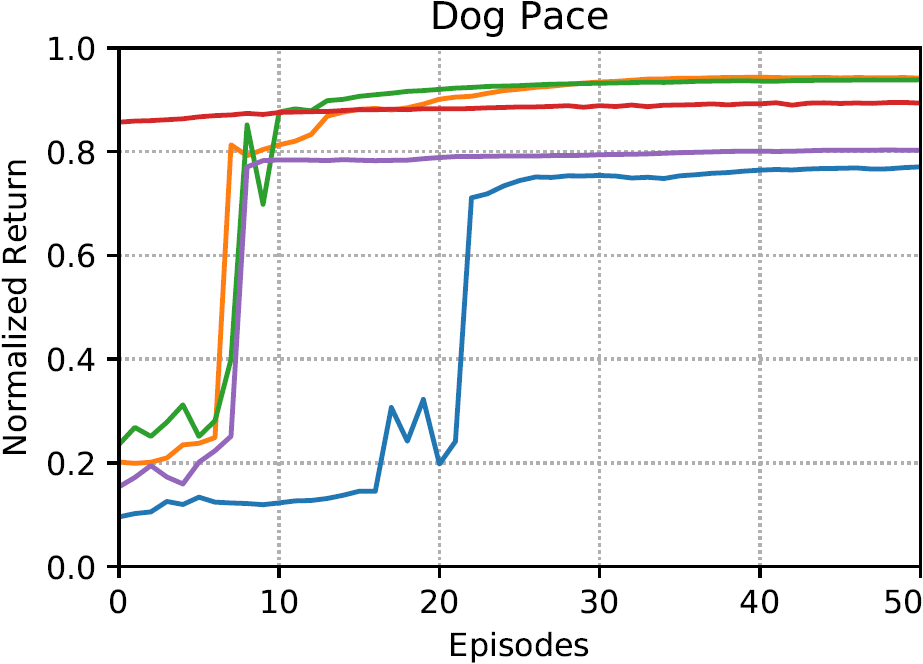}}
    \hfill
    \subfigure{\includegraphics[height=0.35\columnwidth]{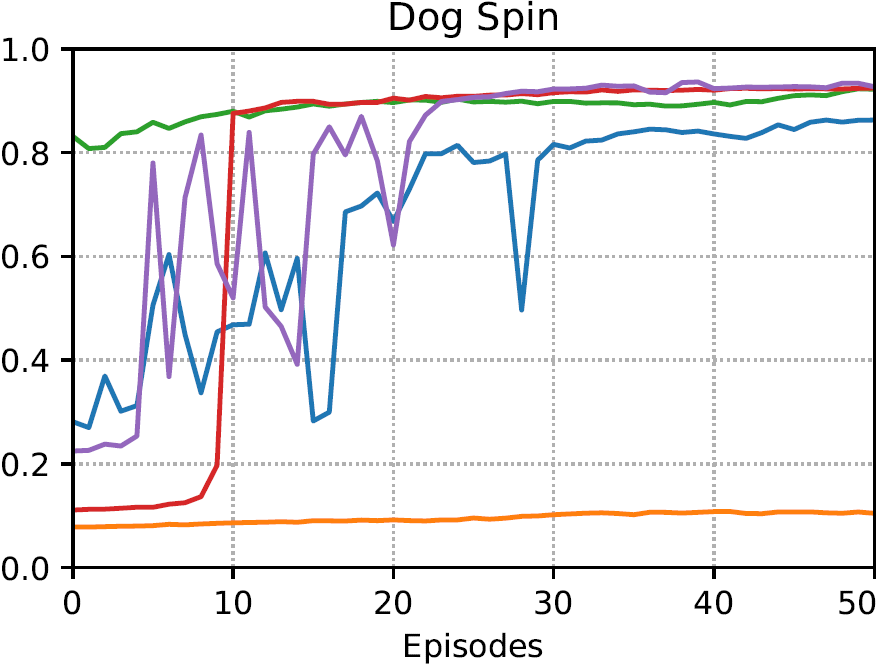}}\\
    \vspace{-0.1cm}
\caption{Learning curves of adapting policies to different  simulated environments using the learned latent space. The policies are able to adapt to new environments in a relatively small number of episodes.}
\label{fig:finetuneCurves}
\vspace{-0.6cm}
\end{figure}

To evaluate the policies' abilities to cope with unfamiliar dynamics, we test the policies in out-of-distribution simulated environments,
where the dynamics parameters are sampled from a larger range of values than those used during training. The range of values used during training and testing are detailed in Table~\ref{tab:dynamicsParams}. Figure~\ref{fig:adaptationCurves} visualizes the performance of the policies in 100 simulated environments with different dynamics. The vertical axis represents the normalized return, and the horizontal axis records the portion of environments in which a policy achieves a return higher than a particular value. For example, in the case of Dog Pace, the adaptive policies achieve a return higher than 0.6 in $50\%$ of the environments, while the robust policy achieves a return higher than 0.6 in $38\%$ of the environments. The experiments are repeated 3 times for each method using policies initialized with different random seeds. In these experiments, the adaptive policies tend to outperform their non-adaptive counterparts across the various skills. This suggests that the adaptation process is able to better generalize to environments that differ from those encountered during training. To analyze the performance of policies during the adaptation process, we record the performance of individual policies after each update iteration. Figure~\ref{fig:finetuneCurves} illustrates the learning curves in 5 different environments for each skill. The policies are generally able to adapt to new environments in a relatively few number of episodes.

\begin{figure}[t]
	\centering
    \subfigure{\includegraphics[height=0.35\columnwidth]{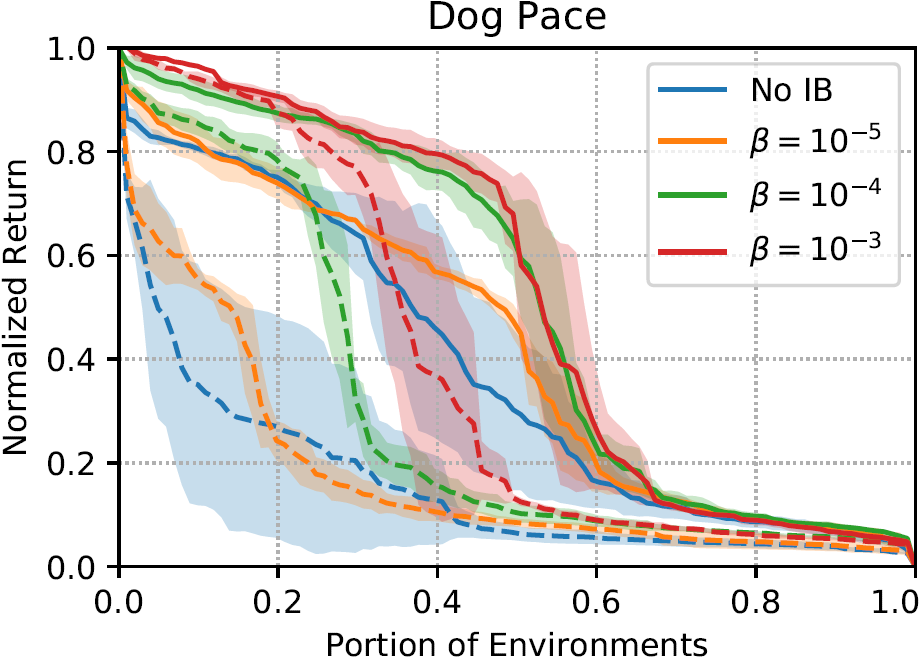}}
    \subfigure{\includegraphics[height=0.35\columnwidth]{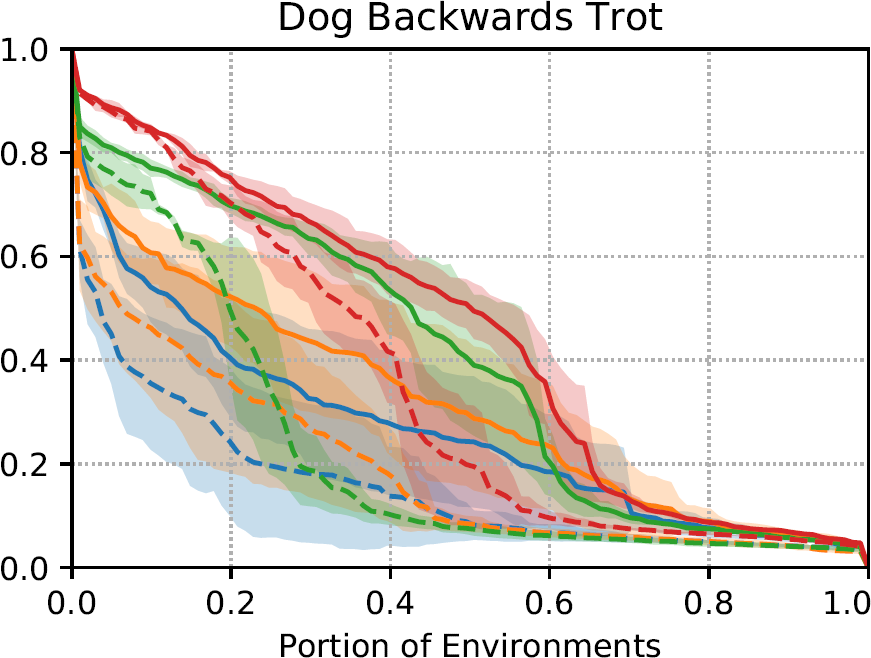}}\\
    \vspace{-0.15cm}
    \subfigure{\includegraphics[height=0.35\columnwidth]{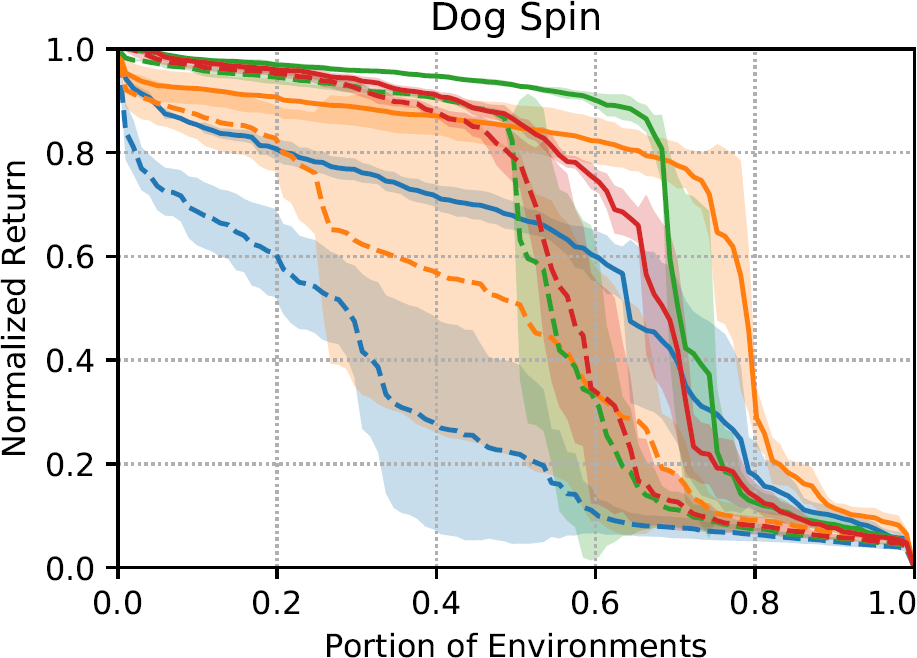}}
    \subfigure{\includegraphics[height=0.35\columnwidth]{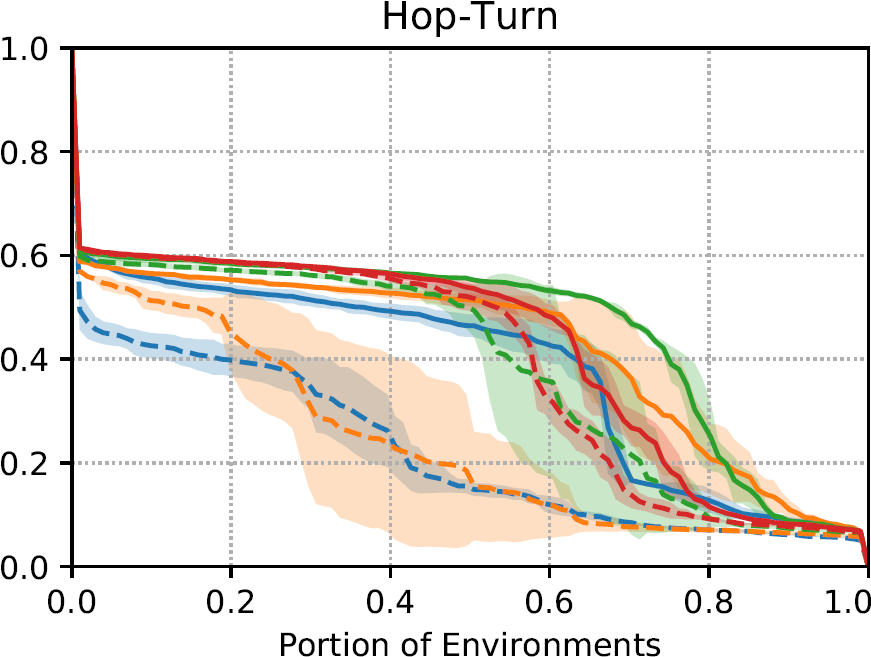}}
    \vspace{-0.2cm}
\caption{Performance of adaptive policies trained with different coefficients $\beta$ for the information penalty. "No IB" corresponds to policies trained without an information bottleneck. The dotted lines represent performance before adaptation, and the solid lines represent after adaptation. 
}
\label{fig:adaptationIBCurves}
\vspace{-0.5cm}
\end{figure}

\subsection{Information Bottleneck}
\vspace{-0.1cm}

Next we evaluate the effects of the information bottleneck on adaptation performance. Figure~\ref{fig:adaptationCurves} summarizes the performance of policies trained with different values of $\beta$ for the information penalty. Larger values of $\beta$ produce policies that access fewer number of bits of information from the dynamics parameters during pre-training. This encourages a policy to be less reliant on precise knowledge of the underlying dynamics, which in turn results in more robust behaviors that attain higher performance before adaptation. However, since the policy's behavior is less dependent on the latent variables, this can also result in less adaptable policies, which exhibit smaller performance improvements after adaptation. Similarly, smaller values of $\beta$ tend to produce less robust but more adaptive policies, exhibiting lower performance before adaptation, but a larger improvement after adaptation. In our experiments, we find that $\beta = 10^{-4}$ provides a good trade-off between robustness and adaptability. We also compare the information-constrained latent representations to the unconstrained counterparts (No IB). The information-constrained policies generally achieve better performance both before and after adaptation.

\section{Discussion and Future Work}
\vspace{-0.1cm}
We presented a framework for learning agile legged-locomotion skills by imitating reference motion data. By simply providing the system with different reference motions, we are able to learn policies for a diverse set of behaviors with a quadruped robot, which can then be efficiently transferred from simulation to the real world. However, due to hardware and algorithmic limitations, we have not been able to learn more dynamic behaviors such as large jumps and runs. Exploring techniques that are able to reproduce these behaviors in the real world could significantly increase the agility of legged robots. The behaviors learned by our policies are currently not as stable as the best manually-designed controllers. Improving the robustness of these learned controllers would be valuable for more complex real-world applications. We are also interested in learning from other sources of motion data, such video clips, which could substantially increase the volume of behavioral data that robots can learn from.

\vspace{-0.1cm}
\section*{Acknowledgements}
\vspace{-0.1cm}
We would like to thank Julian Ibarz, Byron David, Thinh Nguyen, Gus Kouretas, Krista Reymann, Bonny Ho, and the Google Robotics team for their contributions to this work.

\bibliographystyle{plainnat}
\bibliography{references}

\begin{thebibliography}{68}
\providecommand{\natexlab}[1]{#1}
\providecommand{\url}[1]{\texttt{#1}}
\expandafter\ifx\csname urlstyle\endcsname\relax
  \providecommand{\doi}[1]{doi: #1}\else
  \providecommand{\doi}{doi: \begingroup \urlstyle{rm}\Url}\fi

\bibitem[Alemi et~al.(2016)Alemi, Fischer, Dillon, and Murphy]{AlemiFD016}
Alexander~A. Alemi, Ian Fischer, Joshua~V. Dillon, and Kevin Murphy.
\newblock Deep variational information bottleneck.
\newblock \emph{CoRR}, abs/1612.00410, 2016.
\newblock URL \url{http://arxiv.org/abs/1612.00410}.

\bibitem[Apgar et~al.(2018)Apgar, Clary, Green, Fern, and Hurst]{Cassie18}
Taylor Apgar, Patrick Clary, Kevin Green, Alan Fern, and Jonathan Hurst.
\newblock Fast online trajectory optimization for the bipedal robot cassie.
\newblock 06 2018.
\newblock \doi{10.15607/RSS.2018.XIV.054}.

\bibitem[Bledt et~al.(2018)Bledt, Powell, Katz, Carlo, Wensing, and
  Kim]{Bledt2018MITC3}
Gerardo Bledt, Matthew~J. Powell, Benjamin Katz, Jared~Di Carlo, Patrick~M.
  Wensing, and Sangbae Kim.
\newblock Mit cheetah 3: Design and control of a robust, dynamic quadruped
  robot.
\newblock \emph{2018 IEEE/RSJ International Conference on Intelligent Robots
  and Systems (IROS)}, pages 2245--2252, 2018.

\bibitem[Butterworth et~al.(1930)]{butterworth1930theory}
Stephen Butterworth et~al.
\newblock On the theory of filter amplifiers.
\newblock \emph{Wireless Engineer}, 7\penalty0 (6):\penalty0 536--541, 1930.

\bibitem[Chebotar et~al.(2018)Chebotar, Handa, Makoviychuk, Macklin, Issac,
  Ratliff, and Fox]{Chebotar18}
Yevgen Chebotar, Ankur Handa, Viktor Makoviychuk, Miles Macklin, Jan Issac,
  Nathan~D. Ratliff, and Dieter Fox.
\newblock Closing the sim-to-real loop: Adapting simulation randomization with
  real world experience.
\newblock \emph{CoRR}, abs/1810.05687, 2018.
\newblock URL \url{http://arxiv.org/abs/1810.05687}.

\bibitem[Clavera et~al.(2019)Clavera, Nagabandi, Liu, Fearing, Abbeel, Levine,
  and Finn]{clavera2018learning}
Ignasi Clavera, Anusha Nagabandi, Simin Liu, Ronald~S. Fearing, Pieter Abbeel,
  Sergey Levine, and Chelsea Finn.
\newblock Learning to adapt in dynamic, real-world environments through
  meta-reinforcement learning.
\newblock In \emph{International Conference on Learning Representations}, 2019.
\newblock URL \url{https://openreview.net/forum?id=HyztsoC5Y7}.

\bibitem[Coros et~al.(2009)Coros, Beaudoin, and van~de Panne]{Coros09}
Stelian Coros, Philippe Beaudoin, and Michiel van~de Panne.
\newblock Robust task-based control policies for physics-based characters.
\newblock \emph{ACM Trans. Graph. (Proc. SIGGRAPH Asia)}, 28\penalty0
  (5):\penalty0 Article 170, 2009.

\bibitem[Coros et~al.(2010)Coros, Beaudoin, and van~de Panne]{2010-TOG-gbwc}
Stelian Coros, Philippe Beaudoin, and Michiel van~de Panne.
\newblock Generalized biped walking control.
\newblock \emph{ACM Transctions on Graphics}, 29\penalty0 (4):\penalty0 Article
  130, 2010.

\bibitem[Coros et~al.(2011)Coros, Karpathy, Jones, Reveret, and van~de
  Panne]{2011-TOG-quadruped}
Stelian Coros, Andrej Karpathy, Ben Jones, Lionel Reveret, and Michiel van~de
  Panne.
\newblock Locomotion skills for simulated quadrupeds.
\newblock \emph{ACM Transactions on Graphics}, 30\penalty0 (4):\penalty0
  Article TBD, 2011.

\bibitem[Coumans and Bai(2016--2019)]{coumans2019}
Erwin Coumans and Yunfei Bai.
\newblock Pybullet, a python module for physics simulation for games, robotics
  and machine learning.
\newblock \url{http://pybullet.org}, 2016--2019.

\bibitem[de~Lasa et~al.(2010)de~Lasa, Mordatch, and Hertzmann]{delasa2010}
Martin de~Lasa, Igor Mordatch, and Aaron Hertzmann.
\newblock Feature-{B}ased {L}ocomotion {C}ontrollers.
\newblock \emph{ACM Transactions on Graphics}, 29\penalty0 (3), 2010.

\bibitem[Di~Carlo et~al.(2018)Di~Carlo, Wensing, Katz, Bledt, and
  Kim]{di2018dynamic}
Jared Di~Carlo, Patrick~M Wensing, Benjamin Katz, Gerardo Bledt, and Sangbae
  Kim.
\newblock Dynamic locomotion in the {MIT} cheetah 3 through convex
  model-predictive control.
\newblock In \emph{2018 IEEE/RSJ International Conference on Intelligent Robots
  and Systems (IROS)}, pages 1--9. IEEE, 2018.

\bibitem[Duan et~al.(2016)Duan, Schulman, Chen, Bartlett, Sutskever, and
  Abbeel]{DuanSCBSA16}
Yan Duan, John Schulman, Xi~Chen, Peter~L. Bartlett, Ilya Sutskever, and Pieter
  Abbeel.
\newblock Rl{\textdollar}{\^{}}2{\textdollar}: Fast reinforcement learning via
  slow reinforcement learning.
\newblock \emph{CoRR}, abs/1611.02779, 2016.
\newblock URL \url{http://arxiv.org/abs/1611.02779}.

\bibitem[Endo et~al.(2005)Endo, Morimoto, Matsubara, Nakanishi, and
  Cheng]{BipedEndo2005}
Gen Endo, Jun Morimoto, Takamitsu Matsubara, Jun Nakanishi, and Gordon Cheng.
\newblock Learning cpg sensory feedback with policy gradient for biped
  locomotion for a full-body humanoid.
\newblock In \emph{Proceedings of the 20th National Conference on Artificial
  Intelligence - Volume 3}, AAAI’05, page 1267–1273. AAAI Press, 2005.
\newblock ISBN 157735236x.

\bibitem[Featherstone(2007)]{Featherstone2007}
Roy Featherstone.
\newblock \emph{Rigid Body Dynamics Algorithms}.
\newblock Springer-Verlag, Berlin, Heidelberg, 2007.
\newblock ISBN 0387743146.

\bibitem[Finn et~al.(2017)Finn, Abbeel, and Levine]{maml17}
Chelsea Finn, Pieter Abbeel, and Sergey Levine.
\newblock Model-agnostic meta-learning for fast adaptation of deep networks.
\newblock In Doina Precup and Yee~Whye Teh, editors, \emph{Proceedings of the
  34th International Conference on Machine Learning}, volume~70 of
  \emph{Proceedings of Machine Learning Research}, pages 1126--1135,
  International Convention Centre, Sydney, Australia, 06--11 Aug 2017. PMLR.
\newblock URL \url{http://proceedings.mlr.press/v70/finn17a.html}.

\bibitem[Gehring et~al.(2016)Gehring, Coros, Hutler, Bellicoso, Heijnen,
  Diethelm, Bloesch, Fankhauser, Hwangbo, Hoepflinger, and
  Siegwart]{Gehring2016}
Christian Gehring, Stelian Coros, Marco Hutler, Dario Bellicoso, Huub Heijnen,
  Remo Diethelm, Michael Bloesch, Péter Fankhauser, Jemin Hwangbo, Mark
  Hoepflinger, and Roland Siegwart.
\newblock Practice makes perfect: An optimization-based approach to controlling
  agile motions for a quadruped robot.
\newblock \emph{IEEE Robotics \& Automation Magazine}, pages 1--1, 02 2016.
\newblock \doi{10.1109/MRA.2015.2505910}.

\bibitem[Geyer et~al.(2003)Geyer, Seyfarth, and Blickhan]{Geyer03}
Hartmut Geyer, Andre Seyfarth, and Reinhard Blickhan.
\newblock Positive force feedback in bouncing gaits?
\newblock \emph{Proceedings. Biological sciences / The Royal Society},
  270:\penalty0 2173--83, 11 2003.
\newblock \doi{10.1098/rspb.2003.2454}.

\bibitem[Gleicher(1998)]{Gleicher1998RMN}
Michael Gleicher.
\newblock Retargetting motion to new characters.
\newblock In \emph{Proceedings of the 25th Annual Conference on Computer
  Graphics and Interactive Techniques}, SIGGRAPH '98, pages 33--42, New York,
  NY, USA, 1998. ACM.
\newblock ISBN 0-89791-999-8.
\newblock \doi{10.1145/280814.280820}.
\newblock URL \url{http://doi.acm.org/10.1145/280814.280820}.

\bibitem[{Goswami}(1999)]{Goswami1999}
A.~{Goswami}.
\newblock Foot rotation indicator (fri) point: a new gait planning tool to
  evaluate postural stability of biped robots.
\newblock In \emph{Proceedings 1999 IEEE International Conference on Robotics
  and Automation (Cat. No.99CH36288C)}, volume~1, pages 47--52 vol.1, May 1999.
\newblock \doi{10.1109/ROBOT.1999.769929}.

\bibitem[Grimes et~al.(2006)Grimes, Chalodhorn, and Rao]{GrimesCR06}
David~B. Grimes, Rawichote Chalodhorn, and Rajesh P.~N. Rao.
\newblock Dynamic imitation in a humanoid robot through nonparametric
  probabilistic inference.
\newblock In Gaurav~S. Sukhatme, Stefan Schaal, Wolfram Burgard, and Dieter
  Fox, editors, \emph{Robotics: Science and Systems}. The MIT Press, 2006.
\newblock ISBN 0-262-69348-8.
\newblock URL
  \url{http://dblp.uni-trier.de/db/conf/rss/rss2006.html#GrimesCR06}.

\bibitem[Haarnoja et~al.(2018)Haarnoja, Zhou, Ha, Tan, Tucker, and
  Levine]{Haarnoja18}
Tuomas Haarnoja, Aurick Zhou, Sehoon Ha, Jie Tan, George Tucker, and Sergey
  Levine.
\newblock Learning to walk via deep reinforcement learning.
\newblock \emph{CoRR}, abs/1812.11103, 2018.
\newblock URL \url{http://arxiv.org/abs/1812.11103}.

\bibitem[Hanna and Stone(2017)]{AAAI17-Hanna}
Josiah Hanna and Peter Stone.
\newblock Grounded action transformation for robot learning in simulation.
\newblock In \emph{Proceedings of the 31st AAAI Conference on Artificial
  Intelligence (AAAI)}, February 2017.

\bibitem[He et~al.(2018)He, Julian, Heiden, Zhang, Schaal, Lim, Sukhatme, and
  Hausman]{He2018}
Zhanpeng He, Ryan Julian, Eric Heiden, Hejia Zhang, Stefan Schaal, Joseph~J.
  Lim, Gaurav~S. Sukhatme, and Karol Hausman.
\newblock Zero-shot skill composition and simulation-to-real transfer by
  learning task representations.
\newblock \emph{CoRR}, abs/1810.02422, 2018.
\newblock URL \url{http://arxiv.org/abs/1810.02422}.

\bibitem[Heess et~al.(2017)Heess, TB, Sriram, Lemmon, Merel, Wayne, Tassa,
  Erez, Wang, Eslami, Riedmiller, and Silver]{HeessTSLMWTEWER17}
Nicolas Heess, Dhruva TB, Srinivasan Sriram, Jay Lemmon, Josh Merel, Greg
  Wayne, Yuval Tassa, Tom Erez, Ziyu Wang, S.~M.~Ali Eslami, Martin~A.
  Riedmiller, and David Silver.
\newblock Emergence of locomotion behaviours in rich environments.
\newblock \emph{CoRR}, abs/1707.02286, 2017.
\newblock URL \url{http://arxiv.org/abs/1707.02286}.

\bibitem[Hwangbo et~al.(2019)Hwangbo, Lee, Dosovitskiy, Bellicoso, Tsounis,
  Koltun, and Hutter]{Hwangboeaau5872}
Jemin Hwangbo, Joonho Lee, Alexey Dosovitskiy, Dario Bellicoso, Vassilios
  Tsounis, Vladlen Koltun, and Marco Hutter.
\newblock Learning agile and dynamic motor skills for legged robots.
\newblock \emph{Science Robotics}, 4\penalty0 (26), 2019.
\newblock \doi{10.1126/scirobotics.aau5872}.
\newblock URL \url{https://robotics.sciencemag.org/content/4/26/eaau5872}.

\bibitem[{Kim} et~al.(2009){Kim}, {Kim}, {You}, and {Oh}]{KimHumanoid09}
S.~{Kim}, C.~{Kim}, B.~{You}, and S.~{Oh}.
\newblock Stable whole-body motion generation for humanoid robots to imitate
  human motions.
\newblock In \emph{2009 IEEE/RSJ International Conference on Intelligent Robots
  and Systems}, pages 2518--2524, Oct 2009.
\newblock \doi{10.1109/IROS.2009.5354271}.

\bibitem[Kingma and Ba(2014)]{kingma2014adam}
Diederik~P Kingma and Jimmy Ba.
\newblock Adam: A method for stochastic optimization.
\newblock \emph{arXiv preprint arXiv:1412.6980}, 2014.

\bibitem[Kingma and Welling(2013)]{KingmaW13}
Diederik~P. Kingma and Max Welling.
\newblock Auto-encoding variational bayes.
\newblock \emph{CoRR}, abs/1312.6114, 2013.
\newblock URL
  \url{http://dblp.uni-trier.de/db/journals/corr/corr1312.html#KingmaW13}.

\bibitem[Koenemann et~al.(2014)Koenemann, Burget, and
  Bennewitz]{Koenemann2014RealtimeIO}
Jonas Koenemann, Felix Burget, and Maren Bennewitz.
\newblock Real-time imitation of human whole-body motions by humanoids.
\newblock \emph{2014 IEEE International Conference on Robotics and Automation
  (ICRA)}, pages 2806--2812, 2014.

\bibitem[Kohl and Stone(2004)]{KohlS04}
Nate Kohl and Peter Stone.
\newblock Policy gradient reinforcement learning for fast quadrupedal
  locomotion.
\newblock In \emph{ICRA}, pages 2619--2624. IEEE, 2004.
\newblock URL
  \url{http://dblp.uni-trier.de/db/conf/icra/icra2004-3.html#KohlS04}.

\bibitem[Lee et~al.(2019)Lee, Park, Lee, and Lee]{MuscleConLee2019}
Seunghwan Lee, Moonseok Park, Kyoungmin Lee, and Jehee Lee.
\newblock Scalable muscle-actuated human simulation and control.
\newblock \emph{ACM Trans. Graph.}, 38\penalty0 (4), July 2019.
\newblock ISSN 0730-0301.
\newblock \doi{10.1145/3306346.3322972}.
\newblock URL \url{https://doi.org/10.1145/3306346.3322972}.

\bibitem[Lee et~al.(2010)Lee, Kim, and Lee]{BipedLee2010}
Yoonsang Lee, Sungeun Kim, and Jehee Lee.
\newblock Data-driven biped control.
\newblock \emph{ACM Trans. Graph.}, 29\penalty0 (4), July 2010.
\newblock ISSN 0730-0301.
\newblock \doi{10.1145/1778765.1781155}.
\newblock URL \url{https://doi.org/10.1145/1778765.1781155}.

\bibitem[Liu and Hodgins(2018)]{BasketballLiu2018}
Libin Liu and Jessica Hodgins.
\newblock Learning basketball dribbling skills using trajectory optimization
  and deep reinforcement learning.
\newblock \emph{ACM Trans. Graph.}, 37\penalty0 (4), July 2018.
\newblock ISSN 0730-0301.
\newblock \doi{10.1145/3197517.3201315}.
\newblock URL \url{https://doi.org/10.1145/3197517.3201315}.

\bibitem[Liu et~al.(2016)Liu, van~de Panne, and Yin]{2016-TOG-controlGraphs}
Libin Liu, Michiel van~de Panne, and KangKang Yin.
\newblock Guided learning of control graphs for physics-based characters.
\newblock \emph{ACM Transactions on Graphics}, 35\penalty0 (3), 2016.

\bibitem[Lowrey et~al.(2018)Lowrey, Kolev, Dao, Rajeswaran, and
  Todorov]{Lowrey18}
Kendall Lowrey, Svetoslav Kolev, Jeremy Dao, Aravind Rajeswaran, and Emanuel
  Todorov.
\newblock Reinforcement learning for non-prehensile manipulation: Transfer from
  simulation to physical system.
\newblock \emph{CoRR}, abs/1803.10371, 2018.
\newblock URL \url{http://arxiv.org/abs/1803.10371}.

\bibitem[Miura and Shimoyama(1984)]{Miura1984DynamicWO}
Hirofumi Miura and Isao Shimoyama.
\newblock Dynamic walk of a biped.
\newblock \emph{The International Journal of Robotics Research}, 3:\penalty0 60
  -- 74, 1984.

\bibitem[Muico et~al.(2009)Muico, Lee, Popoviundefined, and
  Popoviundefined]{Muico2009}
Uldarico Muico, Yongjoon Lee, Jovan Popoviundefined, and Zoran Popoviundefined.
\newblock Contact-aware nonlinear control of dynamic characters.
\newblock \emph{ACM Trans. Graph.}, 28\penalty0 (3), July 2009.
\newblock ISSN 0730-0301.
\newblock \doi{10.1145/1531326.1531387}.
\newblock URL \url{https://doi.org/10.1145/1531326.1531387}.

\bibitem[{Nakaoka} et~al.(2003){Nakaoka}, {Nakazawa}, {Yokoi}, {Hirukawa}, and
  {Ikeuchi}]{Nakaoka2003}
S.~{Nakaoka}, A.~{Nakazawa}, K.~{Yokoi}, H.~{Hirukawa}, and K.~{Ikeuchi}.
\newblock Generating whole body motions for a biped humanoid robot from
  captured human dances.
\newblock In \emph{2003 IEEE International Conference on Robotics and
  Automation (Cat. No.03CH37422)}, volume~3, pages 3905--3910 vol.3, Sep. 2003.
\newblock \doi{10.1109/ROBOT.2003.1242196}.

\bibitem[Neumann and Peters(2009)]{FQIbyAWR2009}
Gerhard Neumann and Jan~R. Peters.
\newblock Fitted q-iteration by advantage weighted regression.
\newblock In D.~Koller, D.~Schuurmans, Y.~Bengio, and L.~Bottou, editors,
  \emph{Advances in Neural Information Processing Systems 21}, pages
  1177--1184. Curran Associates, Inc., 2009.
\newblock URL
  \url{http://papers.nips.cc/paper/3501-fitted-q-iteration-by-advantage-weighted-regression.pdf}.

\bibitem[OpenAI et~al.(2018)OpenAI, Andrychowicz, Baker, Chociej,
  J{\'{o}}zefowicz, McGrew, Pachocki, Pachocki, Petron, Plappert, Powell, Ray,
  Schneider, Sidor, Tobin, Welinder, Weng, and Zaremba]{Andrychowicz2018}
OpenAI, Marcin Andrychowicz, Bowen Baker, Maciek Chociej, Rafal
  J{\'{o}}zefowicz, Bob McGrew, Jakub~W. Pachocki, Jakub Pachocki, Arthur
  Petron, Matthias Plappert, Glenn Powell, Alex Ray, Jonas Schneider, Szymon
  Sidor, Josh Tobin, Peter Welinder, Lilian Weng, and Wojciech Zaremba.
\newblock Learning dexterous in-hand manipulation.
\newblock \emph{CoRR}, abs/1808.00177, 2018.
\newblock URL \url{http://arxiv.org/abs/1808.00177}.

\bibitem[Peng et~al.(2018{\natexlab{a}})Peng, Andrychowicz, Zaremba, and
  Abbeel]{Sim2Real2018}
X.~B. Peng, M.~Andrychowicz, W.~Zaremba, and P.~Abbeel.
\newblock Sim-to-real transfer of robotic control with dynamics randomization.
\newblock In \emph{2018 IEEE International Conference on Robotics and
  Automation (ICRA)}, pages 1--8, May 2018{\natexlab{a}}.
\newblock \doi{10.1109/ICRA.2018.8460528}.

\bibitem[Peng et~al.(2016)Peng, Berseth, and van~de Panne]{2016-TOG-deepRL}
Xue~Bin Peng, Glen Berseth, and Michiel van~de Panne.
\newblock Terrain-adaptive locomotion skills using deep reinforcement learning.
\newblock \emph{ACM Trans. Graph.}, 35\penalty0 (4):\penalty0 81:1--81:12, July
  2016.
\newblock ISSN 0730-0301.
\newblock \doi{10.1145/2897824.2925881}.
\newblock URL \url{http://doi.acm.org/10.1145/2897824.2925881}.

\bibitem[Peng et~al.(2018{\natexlab{b}})Peng, Abbeel, Levine, and van~de
  Panne]{2018-TOG-deepMimic}
Xue~Bin Peng, Pieter Abbeel, Sergey Levine, and Michiel van~de Panne.
\newblock Deepmimic: Example-guided deep reinforcement learning of
  physics-based character skills.
\newblock \emph{ACM Trans. Graph.}, 37\penalty0 (4):\penalty0 143:1--143:14,
  July 2018{\natexlab{b}}.
\newblock ISSN 0730-0301.
\newblock \doi{10.1145/3197517.3201311}.
\newblock URL \url{http://doi.acm.org/10.1145/3197517.3201311}.

\bibitem[Peng et~al.(2018{\natexlab{c}})Peng, Kanazawa, Malik, Abbeel, and
  Levine]{2018-TOG-SFV}
Xue~Bin Peng, Angjoo Kanazawa, Jitendra Malik, Pieter Abbeel, and Sergey
  Levine.
\newblock Sfv: Reinforcement learning of physical skills from videos.
\newblock \emph{ACM Trans. Graph.}, 37\penalty0 (6), November
  2018{\natexlab{c}}.

\bibitem[Peng et~al.(2019)Peng, Kumar, Zhang, and Levine]{AWRPeng19}
Xue~Bin Peng, Aviral Kumar, Grace Zhang, and Sergey Levine.
\newblock Advantage-weighted regression: Simple and scalable off-policy
  reinforcement learning.
\newblock \emph{CoRR}, abs/1910.00177, 2019.
\newblock URL \url{https://arxiv.org/abs/1910.00177}.

\bibitem[Pinto et~al.(2017)Pinto, Davidson, Sukthankar, and Gupta]{pinto17a}
Lerrel Pinto, James Davidson, Rahul Sukthankar, and Abhinav Gupta.
\newblock Robust adversarial reinforcement learning.
\newblock In Doina Precup and Yee~Whye Teh, editors, \emph{Proceedings of the
  34th International Conference on Machine Learning}, volume~70 of
  \emph{Proceedings of Machine Learning Research}, pages 2817--2826,
  International Convention Centre, Sydney, Australia, 06--11 Aug 2017. PMLR.
\newblock URL \url{http://proceedings.mlr.press/v70/pinto17a.html}.

\bibitem[Pollard et~al.(2002)Pollard, Hodgins, Riley, and Atkeson]{Pollard2002}
Nancy Pollard, Jessica~K. Hodgins, M.J. Riley, and Chris Atkeson.
\newblock Adapting human motion for the control of a humanoid robot.
\newblock In \emph{Proceedings of the IEEE International Conference on Robotics
  and Automation (ICRA '02)}, May 2002.

\bibitem[{Raibert}(1984)]{Hopping84}
M.~H. {Raibert}.
\newblock Hopping in legged systems — modeling and simulation for the
  two-dimensional one-legged case.
\newblock \emph{IEEE Transactions on Systems, Man, and Cybernetics},
  SMC-14\penalty0 (3):\penalty0 451--463, May 1984.
\newblock ISSN 2168-2909.
\newblock \doi{10.1109/TSMC.1984.6313238}.

\bibitem[Raibert(1990)]{raibert1990trotting}
Marc~H Raibert.
\newblock Trotting, pacing and bounding by a quadruped robot.
\newblock \emph{Journal of biomechanics}, 23:\penalty0 79--98, 1990.

\bibitem[Rusu et~al.(2017)Rusu, Večerík, Rothörl, Heess, Pascanu, and
  Hadsell]{pmlr-v78-rusu17a}
Andrei~A. Rusu, Matej Večerík, Thomas Rothörl, Nicolas Heess, Razvan
  Pascanu, and Raia Hadsell.
\newblock Sim-to-real robot learning from pixels with progressive nets.
\newblock In Sergey Levine, Vincent Vanhoucke, and Ken Goldberg, editors,
  \emph{Proceedings of the 1st Annual Conference on Robot Learning}, volume~78
  of \emph{Proceedings of Machine Learning Research}, pages 262--270. PMLR,
  13--15 Nov 2017.
\newblock URL \url{http://proceedings.mlr.press/v78/rusu17a.html}.

\bibitem[Sadeghi and Levine(2016)]{SadeghiL16}
Fereshteh Sadeghi and Sergey Levine.
\newblock Cad2rl: Real single-image flight without a single real image.
\newblock \emph{CoRR}, abs/1611.04201, 2016.
\newblock URL \url{http://arxiv.org/abs/1611.04201}.

\bibitem[Schulman et~al.(2017)Schulman, Wolski, Dhariwal, Radford, and
  Klimov]{PPOSchulmanWDRK17}
John Schulman, Filip Wolski, Prafulla Dhariwal, Alec Radford, and Oleg Klimov.
\newblock Proximal policy optimization algorithms.
\newblock \emph{CoRR}, abs/1707.06347, 2017.
\newblock URL \url{http://arxiv.org/abs/1707.06347}.

\bibitem[Schwind and Koditschek(1998)]{Schwind1998SpringLI}
William~J. Schwind and Daniel~E. Koditschek.
\newblock Spring loaded inverted pendulum running: a plant model.
\newblock 1998.

\bibitem[{Suleiman} et~al.(2008){Suleiman}, {Yoshida}, {Kanehiro}, {Laumond},
  and {Monin}]{Suleiman08}
W.~{Suleiman}, E.~{Yoshida}, F.~{Kanehiro}, J.~{Laumond}, and A.~{Monin}.
\newblock On human motion imitation by humanoid robot.
\newblock In \emph{2008 IEEE International Conference on Robotics and
  Automation}, pages 2697--2704, May 2008.
\newblock \doi{10.1109/ROBOT.2008.4543619}.

\bibitem[Sutton and Barto(1998)]{Sutton1998IRL}
Richard~S. Sutton and Andrew~G. Barto.
\newblock \emph{Introduction to Reinforcement Learning}.
\newblock MIT Press, Cambridge, MA, USA, 1st edition, 1998.
\newblock ISBN 0262193981.

\bibitem[{Tan} et~al.(2016){Tan}, {Xie}, {Boots}, and {Liu}]{Sim2RealTan2016}
J.~{Tan}, Z.~{Xie}, B.~{Boots}, and C.~K. {Liu}.
\newblock Simulation-based design of dynamic controllers for humanoid
  balancing.
\newblock In \emph{2016 IEEE/RSJ International Conference on Intelligent Robots
  and Systems (IROS)}, pages 2729--2736, Oct 2016.
\newblock \doi{10.1109/IROS.2016.7759424}.

\bibitem[Tan et~al.(2018)Tan, Zhang, Coumans, Iscen, Bai, Hafner, Bohez, and
  Vanhoucke]{Tan-RSS-18}
Jie Tan, Tingnan Zhang, Erwin Coumans, Atil Iscen, Yunfei Bai, Danijar Hafner,
  Steven Bohez, and Vincent Vanhoucke.
\newblock Sim-to-real: Learning agile locomotion for quadruped robots.
\newblock In \emph{Proceedings of Robotics: Science and Systems}, Pittsburgh,
  Pennsylvania, June 2018.
\newblock \doi{10.15607/RSS.2018.XIV.010}.

\bibitem[Tedrake et~al.(2004)Tedrake, Zhang, and Seung]{Tedrake2004}
Russ Tedrake, Teresa~Weirui Zhang, and H.~Sebastian Seung.
\newblock Stochastic policy gradient reinforcement learning on a simple 3d
  biped.
\newblock In \emph{Proceedings of the 2004 IEEE/RSJ International Conference on
  Intelligent Robots and Systems (IROS 2004)}, volume~3, pages 2849--2854,
  Piscataway, NJ, USA, 2004. IEEE.
\newblock ISBN 0-7803-8463-6.
\newblock URL \url{http://www.cs.cmu.edu/~cga/legs/01389841.pdf}.

\bibitem[Tobin et~al.(2017)Tobin, Fong, Ray, Schneider, Zaremba, and
  Abbeel]{TobinFRSZA17}
Joshua Tobin, Rachel Fong, Alex Ray, Jonas Schneider, Wojciech Zaremba, and
  Pieter Abbeel.
\newblock Domain randomization for transferring deep neural networks from
  simulation to the real world.
\newblock \emph{CoRR}, abs/1703.06907, 2017.
\newblock URL \url{http://arxiv.org/abs/1703.06907}.

\bibitem[Wang(2018)]{laikago}
Xingxing Wang.
\newblock Laikago {P}ro, {U}nitree {R}obotics, 2018.
\newblock URL
  \url{http://www.unitree.cc/e/action/ShowInfo.php?classid=6&id=355}.

\bibitem[Xie et~al.(2019)Xie, Clary, Dao, Morais, Hurst, and van~de
  Panne]{2019-CORL-cassie}
Zhaoming Xie, Patrick Clary, Jeremy Dao, Pedro Morais, Jonathan Hurst, and
  Michiel van~de Panne.
\newblock Learning locomotion skills for cassie: Iterative design and
  sim-to-real.
\newblock In \emph{Proc. Conference on Robot Learning (CORL 2019)}, 2019.

\bibitem[{Yamane} et~al.(2010){Yamane}, {Anderson}, and {Hodgins}]{Yamane2010}
K.~{Yamane}, S.~O. {Anderson}, and J.~K. {Hodgins}.
\newblock Controlling humanoid robots with human motion data: Experimental
  validation.
\newblock In \emph{2010 10th IEEE-RAS International Conference on Humanoid
  Robots}, pages 504--510, Dec 2010.
\newblock \doi{10.1109/ICHR.2010.5686312}.

\bibitem[Yin et~al.(2007)Yin, Loken, and van~de Panne]{Yin07}
KangKang Yin, Kevin Loken, and Michiel van~de Panne.
\newblock Simbicon: Simple biped locomotion control.
\newblock \emph{ACM Trans. Graph.}, 26\penalty0 (3):\penalty0 Article 105,
  2007.

\bibitem[Yu et~al.(2019{\natexlab{a}})Yu, Kumar, Turk, and Liu]{BipedYu19}
Wenhao Yu, Visak C.~V. Kumar, Greg Turk, and C.~Karen Liu.
\newblock Sim-to-real transfer for biped locomotion.
\newblock \emph{CoRR}, abs/1903.01390, 2019{\natexlab{a}}.
\newblock URL \url{http://arxiv.org/abs/1903.01390}.

\bibitem[Yu et~al.(2019{\natexlab{b}})Yu, Liu, and Turk]{yu2018policy}
Wenhao Yu, C.~Karen Liu, and Greg Turk.
\newblock Policy transfer with strategy optimization.
\newblock In \emph{International Conference on Learning Representations},
  2019{\natexlab{b}}.
\newblock URL \url{https://openreview.net/forum?id=H1g6osRcFQ}.

\bibitem[Yu et~al.(2019{\natexlab{c}})Yu, Tan, Bai, Coumans, and
  Ha]{yu2019learning}
Wenhao Yu, Jie Tan, Yunfei Bai, Erwin Coumans, and Sehoon Ha.
\newblock Learning fast adaptation with meta strategy optimization,
  2019{\natexlab{c}}.

\bibitem[Zhang et~al.(2018)Zhang, Starke, Komura, and Saito]{Zhang2018MNN}
He~Zhang, Sebastian Starke, Taku Komura, and Jun Saito.
\newblock Mode-adaptive neural networks for quadruped motion control.
\newblock \emph{ACM Trans. Graph.}, 37\penalty0 (4):\penalty0 145:1--145:11,
  July 2018.
\newblock ISSN 0730-0301.
\newblock \doi{10.1145/3197517.3201366}.
\newblock URL \url{http://doi.acm.org/10.1145/3197517.3201366}.

\end{thebibliography}

\clearpage

\onecolumn

\appendix

\subsection{Hyperparameters}
\label{app:hyperParams}

Table~\ref{tab:suppHyperparamsPPO} summarizes the hyper-parameter settings for training with proximal-policy optimization (PPO) in simulation, and Table~\ref{tab:suppHyperparamsAWR} shows the hyper-parameters for domain adaptation with advantage-weighted regression (AWR). Gradient descent descent updates are performed using Adam \citep{kingma2014adam}.

\begin{table}[h!]
\centering
    \begin{minipage}{.4\textwidth}
      \centering
      \begin{tabular}{|l|c|}
        \hline
        {\bf Parameter} & {\bf Value} \\ \hline
        {Discount factor $\gamma$} & $0.95$ \\ \hline 
        {Policy Adam learning rate} & $2 \times 10^{-5}$\\ \hline
        {Value function Adam learning rate} & $10^{-5}$ \\ \hline
        {PPO clip threshold} & 0.2  \\ \hline
        {PPO batch size} & 10000 \\ \hline
        {PPO epochs} & 10 \\ \hline
        {Information penalty coefficient ($\beta)$} & $10^{-4}$  \\ \hline
    \end{tabular}
    \caption{Hyper-parameters used during training in simulation with PPO.}
    \label{tab:suppHyperparamsPPO}
    \end{minipage}
    \hspace{0.5cm}
    \begin{minipage}{.4\textwidth}
      \centering
      \begin{tabular}{|l|c|}
        \hline
        {\bf Parameter} & {\bf Value} \\ \hline
        {Discount factor $\gamma$} & $1.0$ \\ \hline 
        {Adam learning rate} & $5 \times 10^{-3}$\\ \hline
        {Gradient steps per iteration} & $10$\\ \hline
        {AWR temperature $\alpha$} & $0.01$\\ \hline
    \end{tabular}
    \caption{Hyper-parameters used for domain adaptation with AWR in the real world.}
    \label{tab:suppHyperparamsAWR}
    \end{minipage}
  \end{table}

\subsection{Performance Statistics}
\label{app:perfStats}

\begin{figure*}[h!]
\centering
    \includegraphics[width=0.9\textwidth]{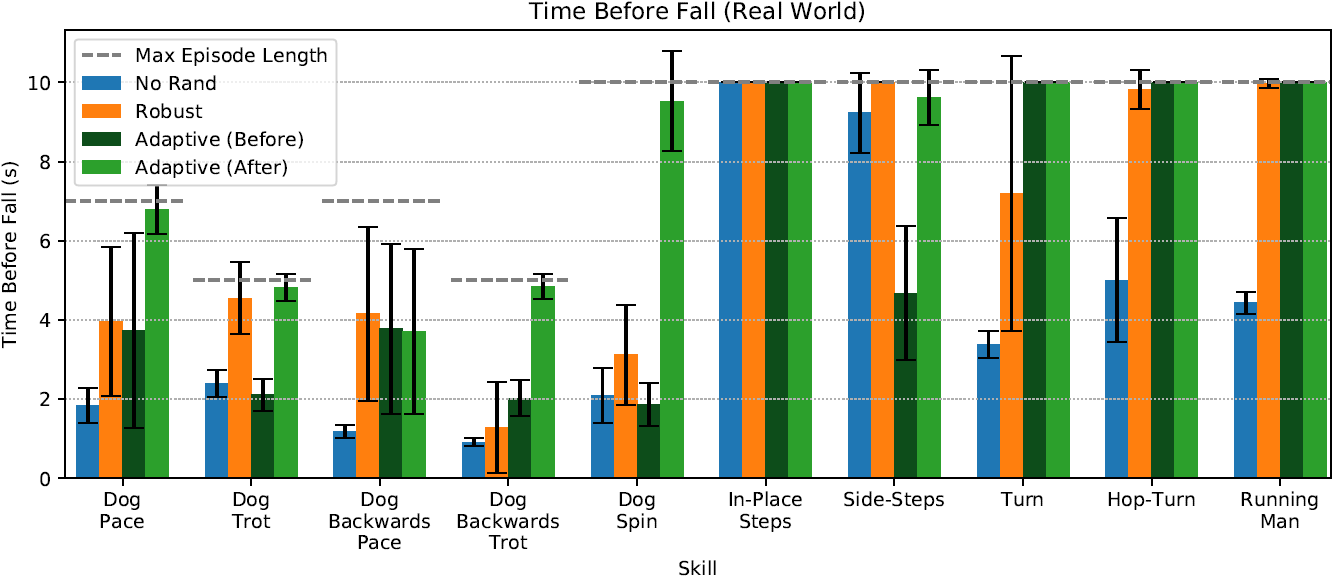}
\caption{Comparison of the time elapsed before the robot falls when deploying various policies in the real world. The adaptive policies are generally able to maintain balance longer than the other baselines policies.}
\label{fig:fallTimeAll}
\end{figure*}

Table~\ref{tab:suppPerfSim} summarizes the performance of the different policies when deployed in the real world, and Table~\ref{tab:suppPerfReal} summarizes performance in simulation. Performance is recorded as the average normalized return, with 0 corresponding to the minimum possible return per episode and 1 being the maximum return. Note that the maximum return may not be achievable, since the reference motions are generally not physically feasible for the robot. Performance is calculated using the average of 3 policies initialized with different random seeds. Performance is evaluated in simulation using the canonical dynamics parameters. Most policies achieve a similar performance in simulation. But when deployed on the real robot, the adaptive policies outperform the non-adaptive policies on most skills. For simpler skills, such as the In-Place Steps and Side-Steps, the robust policy is sufficient for transfer to the real world. But for more dynamic motions, such as Dog Pace and Dog Trot, the robust policy is prone to falling, while the adaptive policies are able to execute the skills more consistently in the real world. Figure~\ref{fig:fallTimeAll} compares the time elapsed before the robot falls under the various policies. Falls are detected when the robot's torso makes contact with the ground. The adaptive policies are often able to maintain balance for a longer period of time than the other methods, with a sizable performance improvement after adaptation. The adaptive policies are often able to reach the maximum episode length without falling.

\begin{table*}[h!]
\centering 
\begin{tabular}{|l|c|c|c|c|}
    \hline
    {\bf Skill} & {\bf No Rand} & {\bf Robust} & {\bf Adaptive (Before)} & {\bf Adaptive (After)} \\ \hline
    {\bf Dog Pace} & $0.128 \pm 0.033$ & $0.350 \pm 0.172$ & $0.395 \pm 0.277$  & $ \cellcolor{gray!30} \bf{0.827 \pm 0.020}$ \\ \hline
    {\bf Dog Trot} & $0.171 \pm 0.031$ & $0.471 \pm 0.102$ &  $0.237 \pm 0.092$ & $\cellcolor{gray!30} \bf{0.593 \pm 0.070}$ \\ \hline
    {\bf Dog Backwards Pace} & $0.067 \pm 0.003$ & $\cellcolor{gray!30} \bf{0.421 \pm 0.244}$ & $0.401 \pm 0.264$ & $0.390 \pm 0.254$ \\ \hline
    {\bf Dog Backwards Trot} & $0.072 \pm 0.004$ & $0.120 \pm 0.126$ & $0.167 \pm 0.048$ & $\cellcolor{gray!30} \bf{0.656 \pm 0.071}$ \\ \hline
    {\bf Dog Spin} & $0.098 \pm 0.033$ & $0.209 \pm 0.081$ & $0.121 \pm 0.035$ & $\cellcolor{gray!30} \bf{0.751 \pm 0.116}$ \\ \hline
    {\bf In-Place Steps} & $0.822 \pm 0.002$ & $\cellcolor{gray!30} \bf{0.845 \pm 0.004}$ & $0.771 \pm 0.001$ & $0.778 \pm 0.002$ \\ \hline
    {\bf Side-Steps} & $0.541 \pm 0.070$ & $\cellcolor{gray!30} \bf{0.782 \pm 0.009}$ & $0.310 \pm 0.114$ & $0.710 \pm 0.057$ \\ \hline
    {\bf Turn} & $0.108 \pm 0.008$ & $0.410 \pm 0.227$ & $0.594 \pm 0.018$ & $\cellcolor{gray!30} \bf{0.606 \pm 0.014}$ \\ \hline
    {\bf Hop-Turn} & $0.174 \pm 0.050$ & $0.478 \pm 0.054$ & $0.493 \pm 0.012$ & $\cellcolor{gray!30} \bf{0.518 \pm 0.005}$ \\ \hline
    {\bf Running Man} & $0.149 \pm 0.004$ & $0.430 \pm 0.031$ & $0.488 \pm 0.045$ & $\cellcolor{gray!30} \bf{0.503 \pm 0.008}$ \\ \hline
\end{tabular}
\caption{Performance statistics of imitating various skills in the real world. Performance is recorded as the average normalized return between [0, 1]. Three policies initialized with different random seeds are trained for each combination of skill and method. The performance of each policy is evaluated over 5 episodes, for a total of 15 trials per method. The method that achieves the highest return for each skill on the real robot is highlighted.}
\label{tab:suppPerfReal}
\end{table*}

\begin{table*}[h!]
\centering 
\begin{tabular}{|l|c|c|c|}
    \hline
    {\bf Skill} & {\bf No Rand} & {\bf Robust} & {\bf Adaptive (Ours)} \\ \hline
    {\bf Dog Pace} & $\cellcolor{gray!30} \bf{0.839 \pm 0.002}$ & $0.820 \pm 0.001$ & $ 0.812 \pm 0.004 $  \\ \hline
    {\bf Dog Trot} & $\cellcolor{gray!30} \bf{0.752 \pm 0.002}$ & $0.727 \pm 0.002$ & $0.718 \pm 0.001$   \\ \hline
    {\bf Dog Backwards Pace} & $\cellcolor{gray!30} \bf{0.843 \pm 0.001}$ & $0.828 \pm 0.001$ & $0.816 \pm 0.002$   \\ \hline
    {\bf Dog Backwards Trot} & $\cellcolor{gray!30} \bf{0.768 \pm 0.002}$ & $0.734 \pm 0.001$ & $0.715 \pm 0.001$   \\ \hline
    {\bf Dog Spin} & $\cellcolor{gray!30} \bf{0.859 \pm 0.001}$ & $0.839 \pm 0.001$ & $0.839 \pm 0.001$  \\ \hline
    {\bf In-Place Steps} & $\cellcolor{gray!30} \bf{0.945 \pm 0.002}$ & $0.938 \pm 0.001$ & $0.935 \pm 0.001$   \\ \hline
    {\bf Side-Steps} & $\cellcolor{gray!30} \bf{0.846 \pm 0.001}$ & $0.808 \pm 0.006$ & $0.820 \pm 0.002$   \\ \hline
    {\bf Turn} & $\cellcolor{gray!30} \bf{0.715 \pm 0.001}$ & $0.666 \pm 0.009$ & $0.675 \pm 0.004$   \\ \hline
    {\bf Hop-Turn} & $\cellcolor{gray!30} \bf{0.628 \pm 0.001}$ & $0.606 \pm 0.005$ & $0.597 \pm 0.001$   \\ \hline
    {\bf Running Man} & $\cellcolor{gray!30} \bf{0.585 \pm 0.003}$ & $0.557 \pm 0.004$ & $0.544 \pm 0.002$   \\ \hline
\end{tabular}
\caption{Performance statistics of imitating various skills in simulation using the canonical dynamics parameters. Performance in simulation is similar across the different methods.}
\label{tab:suppPerfSim}
\end{table*}

\end{document}